\documentclass{article}

\usepackage[margin=1in]{geometry}
\usepackage{graphicx}
\usepackage{subcaption}
\usepackage{amsmath}
\usepackage{amssymb}
\usepackage{booktabs}
\usepackage{multirow}
\usepackage{xcolor}
\usepackage[round]{natbib}
\usepackage[colorlinks=true,linkcolor=blue,citecolor=blue,urlcolor=blue]{hyperref}

\newcommand{\ncffn}{\mbox{NC-FFN}}   
\DeclareMathOperator{\GELU}{GELU}

\title{Explicit Fuzzy Logic in the Feed-Forward Layer\\[2pt]
       {\large Self-Forgetting Quantifiers Discover Legible Grammatical-Licensing Detectors}}
\author{Mark Oskin\\
Professor\\
School of Computer Science and Engineering\\
University of Washington\\
\texttt{mhoskin@uw.edu}}
\date{June 2026}

\begin{document}

\maketitle

\begin{abstract}
The feed-forward (FFN) sublayer is where a transformer materializes the conjunctions and distinctions
its attention has gathered---yet its standard form, a pointwise activation of a linear projection, exposes
no account of \emph{what} it computes. We study a parameter-neutral replacement in which each hidden unit
is an explicit fuzzy set operation on sigmoid-bounded $[0,1]$ membership values: intersection $A\cdot B$
and set-difference $A\cdot(1-B)$, the latter a bounded, \emph{positive} encoding of negation (``$A$ but
not $B$'') that a gated or bilinear unit structurally lacks. We call this a negation-capable FFN
(\ncffn{}). On a controlled reasoning probe ($N$-bit parity in pure feed-forward stacks), bounded
multiplicative units are the most parameter-efficient reasoning basis at shallow depth; at language-model
scale (125M parameters, OpenWebText), \ncffn{} ties the GELU baseline's perplexity at equal parameters,
with every unit carrying explicit logical form. But two limits emerge from the same cause. Genuine
two-operand logic localizes sharply to the embedding-adjacent layer and \emph{erodes} toward soft gating
under language-model training; and the model's one robust grammatical deficit is concentrated in
\emph{licensing} and \emph{quantifiers}---precisely the sequence-level constructions that within-token set
operators cannot express.
Our central result resolves both at once. We add a small, parameter-neutral block of fuzzy
\emph{quantifiers} over the sequence---a soft existential and a soft proportion, each with a per-unit
\emph{learned forgetting rate} initialized at the non-forgetting (sticky) limit. This recovers the grammatical deficit at epoch one---halving the wider gap that opens by epoch two---and modestly leads the baseline on LAMBADA, while making the FFN markedly more legible.
The logical structure now \emph{holds and migrates into depth} rather than eroding; the learned decay
\emph{un-learns its own stickiness}---every unit forgets within a few tokens (median half-life $\sim$$1.5$;
\emph{zero} near-permanent units)---so the quantifier is a local, predictive operator, not a latch; and at
the semantic layers the units read, without dictionary learning, as grammatical \emph{licensing detectors}.
Each fires on a grammatical \emph{licensor} (a comparative, a passive participle, a negative-polarity item)
and its $\sim$$1.5$-token memory carries the membership forward to predict the licensed function word
(\emph{than}, \emph{by}, \emph{nor})---the decay having tuned itself to the grammatical-licensing window.
This legibility is localized (a few percent of the network, concentrated at the late layers) and free only
up to a partition: a \emph{fully} Boolean FFN with no stabilizing GELU fraction still diverges in training,
leaving an end-to-end-legible model an open optimization problem. But within those bounds the result is a
parameter-neutral, language-model-quality transformer that contains a readable,
interpretable-by-construction grammatical mechanism---an explicit account not just of \emph{what} a
feed-forward layer represents but of \emph{how} it licenses.
\end{abstract}

\section{Introduction}
\label{sec:intro}

Mechanistic interpretability has made the most progress where computation is structurally exposed:
attention patterns are legible because attention is, by construction, a weighted routing of values,
and induction or copy circuits can be read off the QK/OV structure. The feed-forward sublayer, which
holds the majority of a transformer's parameters and is widely argued to be where factual and
feature-level computation lives \citep{geva2021kv}, is far less transparent: a pointwise activation of
a linear projection is a dense, polysemantic, sign-agnostic object whose \emph{operation} on its inputs
has no native reading. Post-hoc dictionary methods recover \emph{what} directions a layer represents
\citep{bricken2023monosemanticity,cunningham2023sae}, but not \emph{how} the layer combines them.

We take a complementary, architectural route: choose an FFN whose \emph{combination rule is explicit},
and ask what that costs and reveals. The FFN is a natural home for set logic---it is where attention's
gathered features are conjoined into ``this token is a verb \emph{and} the subject is plural'' or
``a quotation \emph{but not} a heading.'' The last example is the crux. Negation is cheap to
\emph{state} but awkward to \emph{represent}: a raw feature direction has no notion of ``absence,'' so
``not $B$'' has no clean positive form. The fix, known from logical-query embeddings over knowledge
graphs \citep{ren2020beta}, is to make features \emph{bounded} memberships in $[0,1]$, a space closed
under complement, so that ``not $B$'' is simply $1-B$. Embedding the set operators of fuzzy logic---
intersection $A\cdot B$, set-difference $A\cdot(1-B)$, with the complement supplying negation---into the
transformer FFN gives a layer in which every hidden unit is a named \textsc{and}/\textsc{and-not} of two
operands, at no change to the parameter budget, head configuration, or attention. We call it a
negation-capable FFN (\ncffn{}); ``capable'' is an architectural property, not a performance promise.

The operators, the multiplicative-FFN form (GLU and bilinear variants), the operator-softening
optimization pathology, and the negation-via-complement idea are each established prior art
(Section~\ref{sec:related}); our contribution is their assembly inside a transformer language model and
an empirical account of what it does. The account has five parts: four that characterize \ncffn{} itself,
and a fifth that extends it with a sequence-level operator and yields the paper's headline result.

\paragraph{Capability (Section~\ref{sec:capability}).}
The motivating question---does a more \emph{structured} component reason more \emph{compactly}?---is
not answerable from perplexity, which rewards distributional fit, not reasoning. We isolate it with a
controlled probe: $N$-bit parity, the canonical function that is easy for products and hard for sums,
learned by pure feed-forward stacks of each FFN type. Measuring the largest $N$ each solves as a
function of width and depth turns ``reasoning capacity'' into a number. The result is a clean
dissociation: \emph{bounded} multiplicative units (\ncffn{}, sigmoid-bilinear) are the most
parameter-efficient reasoning basis at shallow depth---a single layer outreaches strictly larger
GELU---while \emph{unbounded} bilinear products are useless shallow but compose through depth. The two
properties our architecture toggles, multiplicativity and bounding, turn out to control two different
axes of reasoning efficiency. The small networks here are a controlled instrument for measuring
capability, not a scale claim.

\paragraph{Language modeling (Section~\ref{sec:lm}).}
As a 125M-parameter language model, \ncffn{} is a faithful parameter-neutral drop-in: it ties the GELU
baseline on perplexity (and the tie tightens with training), with a small, persistent deficit on
grammatical structure (BLiMP). We attribute the deficit not to a missing capability but to capacity
\emph{allocation}: a bounded multiplicative unit exposes fewer independent standalone features per
parameter than an unbounded activation, and ordinary language wants standalone features. The capability
the architecture buys is real but \emph{dormant} for next-token prediction, which does not call for it.

\paragraph{Legibility and its dynamics (Section~\ref{sec:interp}).}
Every \ncffn{} unit carries explicit logical form, and a small subset reads as recognizable predicates
with no dictionary learning. Genuine two-operand logic localizes sharply to the embedding-adjacent
layer; the features the readable units track are \emph{causally used} even though the units are
individually redundant. Most strikingly, the network's logical content is \emph{dynamic and
task-shaped}: trained on a task that rewards multiplicative reasoning, the Boolean structure
\emph{crystallizes} precisely at the grokking transition; trained as a language model, which does not
reward it, the same structure \emph{erodes} toward soft gating. The architecture keeps the logic the
task pays for.

\paragraph{A trainability boundary (Section~\ref{sec:trainability}).}
The stabilizing $\GELU$ majority beside the Boolean block is not optional. As the Boolean fraction rises
the model trains longer but then diverges sooner, and a \emph{fully} Boolean FFN diverges within the
first $\sim$$16$k steps; the obvious remedies---bounding the residual write, a parallel linear highway---only
delay it, implicating a saturated-product gradient pathology rather than the write magnitude. Legibility
is thus free only up to a partition, and an end-to-end-legible model is gated on removing this
instability.

\paragraph{Self-forgetting quantifiers and legible licensing (Section~\ref{sec:quantifier}).}
Two of the findings above are limitations with a shared cause: the two-operand logic erodes and localizes
to layer~0, and the grammatical deficit concentrates in \emph{licensing} and \emph{quantifiers}---
constructions about whether a feature occurred \emph{earlier in the sequence}, which \ncffn{}'s
within-token operators cannot express. We add the missing primitive: a parameter-neutral block of fuzzy
quantifiers over the sequence (a soft existential and a soft proportion) with a per-unit \emph{learned
forgetting} rate, initialized at the non-forgetting (sticky) limit. It resolves both limitations at once.
It recovers the grammatical deficit at epoch one---halving the wider epoch-two gap---and leads the baseline on LAMBADA; the logical structure now
\emph{holds and migrates into depth} instead of eroding; the learned decay \emph{un-learns} its sticky
initialization---every unit forgets within a few tokens---so the quantifier is a local, predictive operator
rather than a latch; and at the semantic layers the units read, without dictionary learning, as grammatical
\emph{licensing detectors}, each firing on a grammatical licensor and carrying it forward---on a
$\sim$$1.5$-token memory that the decay has tuned to the licensing window---to predict the licensed function
word. The legibility is localized, not network-wide, but it is a readable, interpretable-by-construction
grammatical mechanism at no parameter or language-model-quality cost.

Together these say that an explicit-combination FFN is a viable, parameter-neutral component that
reasons more compactly on tasks that need it and exposes \emph{how} it computes---at a modest,
well-localized cost; that ``how much logic'' is not a fixed property of the architecture but a readout of
what its training objective rewards; and that giving the feed-forward layer an explicit \emph{sequence}
operator, free to learn how long to remember, turns \ncffn{}'s hardest grammatical failures into its most
legible mechanism---a bank of self-forgetting quantifiers that compute readable grammatical licensing.

\section{Related Work}
\label{sec:related}

\ncffn{} sits at the intersection of several mature lines of work: gated and multiplicative
feed-forward layers, differentiable fuzzy logic, set-operator embeddings for logical query
answering, and the interpretability of transformer feed-forward (FFN) sublayers. The individual
ingredients are all established. Our contribution is one of \emph{placement and lens}: assembling
sigmoid-bounded fuzzy set operators---including an explicit complement/negation term---into a
parameter-neutral FFN sublayer of a decoder-only language model, and characterizing what that
choice does. We organize the prior art accordingly and state, for each line, the precise gap that
remains.

\subsection{Gated and multiplicative feed-forward layers}
Gated Linear Units (GLUs) replace a feed-forward activation with the component-wise product of two
linear projections, one passed through a gate \citep{dauphin2017glu}. \citet{shazeer2020glu}
systematized GLU variants for the transformer FFN---GEGLU, SwiGLU, and a purely \emph{bilinear}
variant $(xW)\odot(xV)$ with no element-wise nonlinearity---and showed they improve language-model
quality at matched parameter count. These are the closest architectural relatives of \ncffn{}: a
GLU is itself a sigmoid-gated multiplicative unit. The distinction is structural. A GLU fuses a
gate and a value into a \emph{single} multiplicatively-gated pathway, whereas \ncffn{} places a
standard activation block \emph{beside} a separate block of explicit set operations and reads them
out through a shared projection; and crucially, no GLU variant carries a complement/negation
($1-B$) term or a set-theoretic reading of its products.

\citet{pearce2024bilinear} and \citet{dooms2024weightdecomp} study the bilinear FFN
$g(x)=(W_1x)\odot(W_2x)$ in transformer language models specifically as a substrate for
\emph{weight-based} mechanistic interpretability: because the layer is a pure bilinear form with no
nonlinearity, each output is a quadratic form whose interaction structure can be recovered by
eigendecomposition. They report that bilinear MLPs are a competitive, parameter-neutral drop-in for
activation MLPs. This is the single nearest piece of prior art to ours, and we treat it as both the
behavioral control and the head-to-head interpretability comparison
(Sections~\ref{sec:methodology},~\ref{sec:interp}). The differences are specific and load-bearing:
bilinear operands are unbounded and signed rather than sigmoid-bounded to $[0,1]$; there is no
complement/negation term; and there is no set-operator semantics. As we show, these are not cosmetic:
the sigmoid bounding gives \ncffn{} a graceful single-operand fallback that a raw bilinear unit
structurally lacks.

Neural Arithmetic Logic Units \citep{trask2018nalu} use a sigmoid gate to select between additive and
multiplicative pathways for systematic numerical extrapolation---adjacent gated-multiplicative
machinery, but aimed at arithmetic rather than set logic and not used as a general FFN.

\subsection{Differentiable fuzzy logic and logic-gate networks}
The operators \ncffn{} uses are textbook continuous (fuzzy) relaxations of Boolean connectives. The
product t-norm realizes conjunction as $A\cdot B$, the strong complement realizes negation as $1-A$,
and set-difference follows as $A\cdot(1-B)$. Neuro-symbolic frameworks such as Logic Tensor Networks
\citep{badreddine2022ltn} and Logical Neural Networks \citep{riegel2020lnn} build differentiable logical
formulae from such operators, and \citet{vankrieken2022fuzzy} analyze their optimization behavior,
showing that gradient descent tends to \emph{soften} crisp fuzzy operators because product-form gates
suffer vanishing gradients as operands saturate. Differentiable logic-gate networks
\citep{petersen2022difflogic,petersen2024convlogic} learn, per neuron, one of the sixteen two-input
Boolean functions via temperature-annealed relaxations, and achieve strong efficiency on vision
benchmarks. These works establish the operators and their optimization pathologies, but always as a
\emph{standalone} logical network (typically for vision or tabular data), never as an FFN sublayer
inside a transformer trained for language modeling. \citet{vankrieken2022fuzzy} in particular
predicts---in non-transformer settings---the softening we observe; our depth-resolved measurements
(Section~\ref{sec:interp}) make this concrete inside a language model and localize where the
softening does \emph{not} happen. Pushed past a roughly even split toward a fully Boolean layer, this
same saturation pathology no longer merely \emph{softens} the operators but destabilizes training
outright (Section~\ref{sec:trainability}).

\subsection{Quantifiers, aggregation, and learned forgetting}
The within-token set operators above have a natural sequence extension---\emph{quantifiers}---and our
self-forgetting quantifier block (Section~\ref{sec:quantifier}) draws on three established lines. First,
\emph{quantifier aggregation} in differentiable logic: fuzzy and neuro-symbolic frameworks realize the
existential as a maximum (or a soft $p$-mean/log-sum-exp) and the universal as a mean or product over a
domain. Logic Tensor Networks \citep{badreddine2022ltn} and Logical Neural Networks \citep{riegel2020lnn}
build exactly these quantifier aggregations, and \citet{vankrieken2022fuzzy} analyze their gradients. Our
existential (a leaky cumulative maximum) and proportion (a running mean) are the causal, left-to-right
specializations of those aggregations over the prefix of the sequence. Second, \emph{learned forgetting}:
a decayed cumulative max or moving average with a \emph{learned} per-unit rate is, mechanistically, the
forget gate of gated recurrent networks \citep{gers2000forget}---a leaky integrator whose time constant is
trained rather than fixed. Our contribution is to give a fuzzy \emph{quantifier} this learned temporal
scope, so the operator can learn its own window and, as we find, \emph{un-learn} a sticky cumulative
initialization toward a short, predictive memory; to our knowledge prior fuzzy quantifiers aggregate over a
fixed or unbounded domain without a learned decay, and none has been placed as an FFN sublayer of a
language model. Third, \emph{grammatical licensing}: the constructions our units detect---NPI licensing,
comparatives, passives---are the long-distance dependencies that targeted grammatical evaluations such as
BLiMP \citep{warstadt2020blimp} probe, and that are inherently about whether a licensor occurred earlier in
the string. Where the interpretability of licensing is otherwise a post-hoc reading of an opaque model, our
quantifier units realize a licensing detector \emph{by construction}.

\subsection{Set operators and negation in embedding space}
A parallel line uses fuzzy/box/cone set operators to answer logical queries over knowledge graphs.
BetaE \citep{ren2020beta} represents entities as Beta distributions and implements
conjunction, disjunction, and---notably---\emph{negation} as a closed operation, motivated by queries
of the form ``$X$ and $Y$ but not $Z$.'' Query2box \citep{ren2020query2box}, ConE
\citep{zhang2021cone}, and FuzzQE \citep{chen2022fuzzqe} develop related set-operator embeddings, the
last using explicit fuzzy logic. These works are the origin of the complement-for-negation idea we
adopt, and their motivating use case is essentially identical to ours at the conceptual level. The gap
is again placement: the operators live in a dedicated \emph{query-answering head} over a knowledge
graph, not as a general-purpose FFN in a language model trained on next-token prediction. An
adversarial citation-forward sweep from this literature did not surface any migration of the operators
into an LM FFN.

\subsection{Interpretability of feed-forward sublayers}
A substantial body of work treats the FFN as the locus of interpretable computation:
\citet{geva2021kv} cast FFN layers as key-value memories, Softmax Linear Units
\citep{elhage2022solu} modify the activation to encourage more interpretable, less polysemantic units,
and Codebook Features \citep{tamkin2023codebook} bind hidden states to a small set of discrete, nameable
codes via a vector-quantization bottleneck---the \emph{interpretable-by-construction} intervention
closest in spirit to ours, though it constrains the activation's discreteness rather than, as \ncffn{}
does, the layer's combination rule.
More broadly, individual neurons are known to be \emph{polysemantic} due to superposition
\citep{elhage2022superposition}, which has motivated dictionary-learning methods---sparse autoencoders---
that recover more monosemantic features post hoc \citep{bricken2023monosemanticity,cunningham2023sae}.
\ncffn{} relates to this line in two ways. First, like \citet{pearce2024bilinear}, it is an
architecture chosen partly for interpretability; but where bilinear interpretability is a global,
weight-based spectral decomposition, \ncffn{} exposes \emph{explicit logical form}---each unit is an
\textsc{and}/\textsc{and-not} of two operands, including a negation a bilinear eigenvector cannot
express. Second, our findings are consistent with superposition: the \emph{operands} of \ncffn{}
units are themselves distributed/polysemantic directions, so the architecture supplies the
combination rule for free but not, in general, monosemantic predicates---a small directly-readable
subset notwithstanding (Section~\ref{sec:interp}).

\subsection{Neuro-fuzzy and fuzzy-membership transformers}
Finally, an applied literature combines fuzzy logic with transformers, largely in forecasting, fault
diagnosis, and control. The nearest work to ours on keywords is
\citet{semanticfusion2025}, which augments a transformer language model with a parallel
fuzzy-\emph{membership} feature channel (interpretable per-token features such as part-of-speech cues,
sentiment polarity, and boundary flags) graded by differentiable membership functions and fused via a
gated adapter with an auxiliary reconstruction loss. This shares the words ``fuzzy,'' ``membership,''
and ``language model,'' but the mechanism is a \emph{side-channel adapter} that injects hand-specified
features, not a set-operator FFN computed from the residual stream. It does not implement set
operators, has no complement/negation term, and does not replace the FFN.

\subsection{Summary}
Taken together, the operators (fuzzy t-norms with complement), the multiplicative-FFN form (GLU,
bilinear), the optimization pathology (operator softening), and the negation-via-complement motivation
(KG query embeddings) are all prior art that we build on. What is, to our knowledge,
novel is the specific assembly: \emph{sigmoid-bounded fuzzy set operators, including an explicit
complement/negation term, as a parameter-neutral feed-forward sublayer of a decoder-only language
model}---and, as the paper's headline extension, a \emph{self-forgetting fuzzy quantifier} over the
sequence (a causal existential/proportion with a per-unit learned forgetting rate) as a sublayer of the
same model, which we find computes legible grammatical licensing. The defensible fingerprint that
distinguishes \ncffn{} from its nearest neighbors is the
explicit $(1-B)$ complement inside a transformer FFN: bilinear MLPs have the multiplicative structure
but no bounding and no complement \citep{shazeer2020glu,pearce2024bilinear}; GLUs have the sigmoid gate
but no set-operator framing \citep{shazeer2020glu}; differentiable logic networks have the operators but
are not transformer LMs \citep{petersen2024convlogic}; the KG-query embeddings have the operators and
negation but in a query head \citep{ren2020beta}; and the nearest fuzzy-membership LM
\citep{semanticfusion2025} is a side-channel adapter rather than an FFN operator. Our aim is therefore
not a new operator but a new view of an old one inside the transformer FFN, and an empirical account of
what that view reveals.

\section{The \ncffn{} Layer}
\label{sec:architecture}

\subsection{Background: the transformer FFN}
A standard transformer feed-forward sublayer maps $x\in\mathbb{R}^{d}$ through an up-projection, an
element-wise activation, and a down-projection:
\begin{equation}
\mathrm{FFN}(x) = W_o\,\GELU(W_{\mathrm{in}}\,x), \qquad W_{\mathrm{in}}\in\mathbb{R}^{d_{\mathit{ff}}\times d},\;\; W_o\in\mathbb{R}^{d\times d_{\mathit{ff}}},
\end{equation}
with $2\,d\,d_{\mathit{ff}}$ parameters. \ncffn{} replaces the activation with explicit fuzzy set
operations while preserving this parameter budget exactly.

\subsection{Fuzzy set operators as hidden units}
We read each hidden unit as a fuzzy set-membership value in $[0,1]$. From the layer input we form two
membership vectors via independent linear projections and a logistic nonlinearity,
\begin{equation}
A = \sigma(W_a\,x),\qquad B = \sigma(W_b\,x),\qquad W_a,W_b\in\mathbb{R}^{m\times d},
\end{equation}
and combine them with product-form fuzzy operators. The default configuration uses intersection and
set-difference,
\begin{equation}
\label{eq:ops}
h_{\cap} = A\odot B \quad(\text{``}A\text{ and }B\text{''}),
\qquad
h_{\setminus} = A\odot(1-B)\quad(\text{``}A\text{ and not }B\text{''}),
\end{equation}
where $\odot$ is the element-wise product. The complement $1-B$ supplies an explicit, bounded,
\emph{positive} encoding of negation: a unit can fire for the presence of one feature and the
\emph{absence} of another. This explicit negation primitive is the architectural feature that
distinguishes the layer from its multiplicative neighbors---a bilinear or gated unit has no clean
\emph{positive} encoding of ``$A$ and not $B$''---and we accordingly call it a \emph{negation-capable}
FFN (\ncffn{}). We stress that ``capable'' denotes an \emph{architectural} property, not a performance
claim. A standard FFN can \emph{approximate} negated functions; what \ncffn{} adds is an explicit,
bounded, interpretable negation primitive. And as Section~\ref{sec:interp} shows, possessing the
primitive does not by itself improve---indeed it slightly degrades---performance on negation-sensitive
grammar: the name denotes the capability, not its benefit.
The combined hidden vector is read out through a single down-projection,
\begin{equation}
\ncffn(x) = W_o\,h,\qquad h = [\,h_{\cap}\,;\,h_{\setminus}\,]\in\mathbb{R}^{2m},\qquad W_o\in\mathbb{R}^{d\times 2m}.
\end{equation}
We refer to a paired $(A_i,B_i)$ and the operations built from it as a \emph{unit}, and to the explicit
\textsc{and}/\textsc{and-not} structure as the unit's logical \emph{form}.

\subsection{Parameter neutrality}
For $k$ operators built from a single $(A,B)$ pair, the input side comprises $W_a,W_b$ ($2m$ rows) and
the output side $W_o$ has $H=km$ columns, giving $(2m + km)\,d = (2+k)\,m\,d$ parameters. A pure
\ncffn{} layer with $k$ operators is therefore parameter-matched to a $\GELU$ FFN when
\begin{equation}
(2+k)\,m = 2\,d_{\mathit{ff}}.
\end{equation}
Thus \ncffn{} does not change the layer's parameter count, head configuration, or attention; it spends
the same budget on constrained multiplicative set operations rather than free, unbounded activations.
Because the bounded operators occupy capacity a $\GELU$ block would otherwise have, the
\emph{effective} flexibility per unit is lower; whether this trades favorably is an empirical question
(Section~\ref{sec:interp}).

\subsection{The hybrid partition}
\label{sec:hybrid}
A pure \ncffn{} layer is difficult to train. As operands saturate, the product-form gradients vanish
(the pathology analyzed by \citet{vankrieken2022fuzzy}), and an all-positive, bounded basis is a poor
substrate for the output projection; in our experiments a pure layer trains normally for thousands of
steps and then diverges abruptly, a conditional instability we characterize in
Section~\ref{sec:trainability}. We
therefore use a \emph{hybrid partition}: a fraction of the layer budget is given to a standard $\GELU$
``gradient highway'' and the remainder to the set-operator block, read out through one shared $W_o$:
\begin{equation}
h = [\,\GELU(W_g\,x)\,;\,h_{\cap}\,;\,h_{\setminus}\,],\qquad W_g\in\mathbb{R}^{g\times d}.
\end{equation}
With $g$ $\GELU$ units and $m$ operator pairs, the input side is $(g+2m)d$ and the output side
$H d=(g+km)d$, so parameter neutrality with a $d_{\mathit{ff}}$ $\GELU$ FFN holds when
\begin{equation}
\label{eq:msize}
m = \frac{2\,d_{\mathit{ff}} - 2g}{2+k}.
\end{equation}
We parameterize the split by the $\GELU$ budget fraction $\rho = g/d_{\mathit{ff}}$; our main model uses
$\rho=0.75$ (a $\GELU$-dominant $3{:}1$ split). Two further details make the partition fair. The Boolean
output columns of $W_o$ are \emph{zero-initialized} (re-zeroed after standard initialization), so the
layer begins as a pure $g$-wide $\GELU$ FFN and the set-operator block is wired in only if training
finds it useful (a function-preserving initialization). And because the bounded Boolean block has a
capped output norm while $\GELU$ is unbounded, we apply a per-block RMS normalization (a single scalar
gain per block, parameter-negligible) before $W_o$ so the two blocks enter the read-out on
weight-decay-fair footing.

\subsection{A property: graceful degeneration}
\label{sec:degeneration}
The sigmoid bounding endows \ncffn{} with a property its bilinear cousin lacks. If the second operand
of a unit becomes uninformative---$W_b\!\to\!0$, hence $B=\sigma(W_b x)\!\to\!\tfrac12$, a
\emph{non-zero} constant---then $h_{\cap},h_{\setminus}\!\to\!\tfrac12 A$, and the unit collapses to a
single-operand, sigmoid-gated feature: an ordinary bounded activation. The Boolean unit can therefore
\emph{relax} into a normal activation MLP wherever two-operand logic is not useful. A raw bilinear unit
$(W_1x)(W_2x)$ has no such fallback: a linear projection has no non-zero constant state, so zeroing one
operand kills the unit, and its only single-operand mode is the unbounded quadratic $(wx)^2$ obtained
when $W_1\!\parallel\!W_2$. This difference---a graceful single-operand fallback present under sigmoid
bounding and absent for raw bilinear---is central to interpreting the depth-resolved behavior we report
in Section~\ref{sec:interp}, and motivates the bilinear control of Section~\ref{sec:methodology}.

\section{Experimental Setup}
\label{sec:methodology}

\subsection{Models}
Our primary model is a 12-layer, GPT-2-small-scale decoder ($d=768$, 12 heads, $d_{\mathit{ff}}=3072$,
sequence length $2048$, $125.1$M parameters). The \ncffn{} model replaces every FFN with the hybrid
layer of Section~\ref{sec:hybrid} at $\rho=0.75$ and operators $\{\cap,\setminus\}$ (intersection and
set-difference), giving, per layer, $g=2304$ $\GELU$ units and $m=384$ operator pairs ($768$ Boolean
hidden units). The parameter-matched baseline (\textsc{gelu}) is an identical architecture with standard
$\GELU$ FFNs. We additionally use a 4-layer variant of each architecture for cheaper ablations.

\subsection{Training}
All models are trained on OpenWebText with the same recipe: batch size $16$, sequence length $2048$,
AdamW ($\beta=(0.9,0.95)$, weight decay $0.1$), peak learning rate $3\times10^{-4}$ with $2000$ warmup
steps and cosine decay to $10\%$ of peak over one epoch ($272{,}687$ steps). We checkpoint weights at
$50$k-step intervals ($50$k--$250$k) in addition to epoch boundaries, which enables the training-trajectory analyses
in Section~\ref{sec:interp}. A second epoch (analyzed in later sections) continues from the
epoch-1 optimizer state with a fresh cyclical-cosine cycle (warmup $500$, same peak and decay),
matching the multi-epoch schedule of the baseline.

\subsection{Evaluation}
We report dev perplexity and zero-shot accuracy on a standard small-model suite via the
\textsc{lm-eval} harness \citep{gao2023lmeval}: LAMBADA \citep{paperno2016lambada}, BLiMP (67 subtasks)
\citep{warstadt2020blimp}, ARC-Easy \citep{clark2018arc}, HellaSwag \citep{zellers2019hellaswag}, and
WinoGrande \citep{sakaguchi2020winogrande}. At this scale ARC-Easy, HellaSwag, and WinoGrande carry little discriminating signal for all models---HellaSwag and WinoGrande sit at or near the random baseline---so we treat LAMBADA (long-range semantic prediction) and BLiMP (grammatical structure, with per-subtask resolution) as the discriminating benchmarks and report the rest for completeness. All evaluations use a right-padded batching adapter; we verified the
adapter against published GPT-2 LAMBADA accuracy.

\subsection{Probing the Boolean block}
We use three complementary, inference-only probes, all run on CPU from saved checkpoints.

\paragraph{Engagement and realization.}
For each layer we capture the FFN input on held-out tokens and recompute the operand activations $A,B$
and the two block contributions to the residual write through the shared $W_o$. We report (i) the
\emph{bool share}---the fraction of the per-token FFN write carried by the Boolean block---as a measure
of how much the model \emph{uses} the operators, and (ii) operand statistics ($\bar A$, $\bar B$, and
$\overline{A\!\cdot\!B}$) as a measure of whether genuine \emph{two-operand} logic is \emph{realized}:
a unit doing real set logic must have both an informative $A$ and an informative $B$, whereas a unit
with $B\!\approx\!0.5$ has degenerated to single-operand gating (Section~\ref{sec:degeneration}).

\paragraph{Causal ablation.}
To measure necessity rather than mere activation, we zero the Boolean output columns of $W_o$ (globally,
or one layer at a time) and measure the perplexity penalty on a fixed, paired held-out batch. Because a
large ablation grows more damaging over training simply through co-adaptation, we include a
\emph{control}: ablating an equal-sized random subset of the $\GELU$ columns, which isolates whether a
growing Boolean ablation penalty is Boolean-specific or generic.

\paragraph{Logic-unit scout.}
To find units doing genuine two-operand logic and read their predicates, we rank unit pairs by two
criteria: $A$ is a \emph{selective} sparse gate (low firing rate) and $B$ is \emph{informative when $A$
fires} (it takes both high and low values on $A$-firing tokens, rather than a constant). For the
resulting units we read the predicate by collecting the tokens that trigger the $A$-gate and then, among
$A$-firing positions, the tokens that $B$ \emph{includes} ($A\!\cdot\!B$) versus \emph{excludes}
($A\!\cdot\!(1-B)$)---rendering the unit's \textsc{and}/\textsc{and-not} statement directly.

\subsection{The bilinear control ladder}
\label{sec:ladder}
Because our central claim is about the \emph{view}---fuzzy set operators as an FFN---rather than a new
operator, the key control is whether \ncffn{} behaves and explains differently from a matched bilinear
FFN, the nearest prior art \citep{pearce2024bilinear}. A bilinear \emph{layer} is itself competitive
with $\GELU$ \citep{shazeer2020glu,pearce2024bilinear}, so it is a fair baseline rather than a weakened
one, and a measured difference is attributable to what we change. We therefore compare a ladder of
parameter-neutral $\rho=0.75$ hybrids that differ \emph{only} in the non-$\GELU$ block: (i) \ncffn{}
$[A\!\cdot\!B,\,A\!\cdot\!(1-B)]$ (bounding $+$ complement); (ii) a \emph{sigmoid-bilinear} block
$\sigma(W_1x)\!\cdot\!\sigma(W_2x)$ (bounding, no complement), which---unlike raw bilinear---shares
\ncffn{}'s graceful degeneration and so isolates the contribution of the complement/set-operator
structure; (iii) a \emph{raw bilinear} block $(W_1x)\!\cdot\!(W_2x)$ (neither), which isolates the
contribution of sigmoid bounding; and (iv) the $\GELU$ anchor. We realize this ladder in pure form as the four FFN types of the capability probe (Section~\ref{sec:capability}), and the same parameter-neutral partition can be instantiated at language-model scale as a control.

\section{Results I: Reasoning Capability}
\label{sec:capability}

Perplexity cannot answer our motivating question---whether a more structured FFN reasons more
compactly---because next-token prediction rewards distributional fit, not reasoning, and a model can be
fluent without composing anything. We therefore isolate capability with a controlled probe and measure
it as a number.

\subsection{A capability probe: parity in pure feed-forward stacks}
\label{sec:parityprobe}
$N$-bit parity is the canonical separator of additive from multiplicative computation. In the
$\pm1$ encoding it is a single product $\prod_i x_i$, trivial for a multiplicative unit to
\emph{represent} but requiring exponentially many threshold units---and offering no benefit from
depth---for a purely additive (e.g.\ ReLU/GELU) one. It is exactly the kind of compositional function a
structured component should handle more cheaply, and it is uncontaminated by the distributional
statistics that let language models fake competence.

To attribute any difference to the FFN alone we remove attention entirely: each model is a pure
feed-forward stack---a linear embedding of the $N$ input bits ($\to\mathbb{R}^{128}$), $L$ residual FFN
blocks, and a linear head---trained directly to predict the parity bit. We sweep the FFN \emph{type},
the depth $L\in\{1,2,3,4\}$, the hidden width $d_{\mathit{ff}}$, and the difficulty $N\in\{1,\dots,12\}$,
with multiple seeds, and summarize each configuration by its \emph{reach}: the largest $N$ it solves
(best accuracy $\ge 0.75$, averaged over seeds). Four FFN types isolate the two properties our
architecture toggles, here in \emph{pure} form (no GELU partition, so the block is entirely the named
operator): \textsc{gelu} (additive); \emph{raw-bilinear} $(W_1x)\!\cdot\!(W_2x)$ (multiplicative,
unbounded); \emph{sigmoid-bilinear} $\sigma(W_1x)\!\cdot\!\sigma(W_2x)$ (multiplicative, bounded, no
complement); and \ncffn{} $[\sigma(W_ax)\sigma(W_bx),\,\sigma(W_ax)(1-\sigma(W_bx))]$ (multiplicative,
bounded, with complement). The networks are deliberately small: they are a measuring instrument for
per-parameter reasoning efficiency, not a scale claim.

\subsection{Bounding and depth control two different axes of reasoning}
\begin{figure}[t]
\centering
\includegraphics[width=0.62\linewidth]{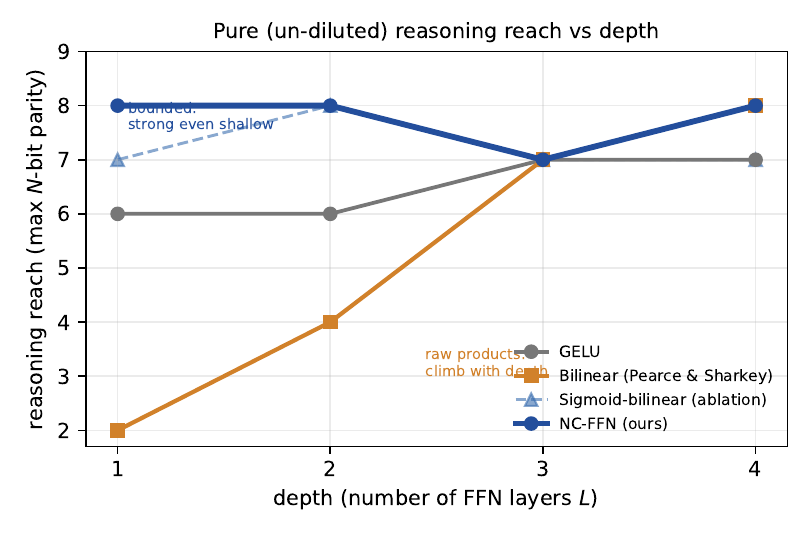}
\caption{Reasoning reach versus depth for the four pure FFN types. The two \emph{bounded} multiplicative
lines (\ncffn{}, sigmoid-bilinear) start high and stay high; \emph{unbounded} raw-bilinear climbs
steeply from a useless-shallow base, tracking the degree-doubling of composed products; the additive
GELU baseline plateaus in between. Bounding buys width-efficient shallow reasoning; unbounding buys
depth-composition.}
\label{fig:depthreach}
\end{figure}

Figure~\ref{fig:depthreach} and Table~\ref{tab:reach} report reach as a function of depth at fixed
width. The picture is a clean dissociation rather than a single ordering.

\begin{table}[t]
\centering
\caption{Reasoning reach (largest $N$-bit parity solved, mean over 5 seeds) versus depth $L$, pure
feed-forward stacks at $d_{\mathit{ff}}=256$. Bounded multiplicative units (\ncffn{}, sigmoid-bilinear)
dominate at shallow depth; unbounded raw-bilinear is useless shallow but climbs steeply with depth;
GELU plateaus between.}
\label{tab:reach}
\begin{tabular}{lcccc}
\toprule
FFN type & $L=1$ & $L=2$ & $L=3$ & $L=4$ \\
\midrule
GELU (additive)                 & 6 & 6 & 7 & 7 \\
Raw-bilinear (mult., unbounded) & 2 & 4 & 7 & 8 \\
Sigmoid-bilinear (mult., bounded) & 7 & 8 & 7 & 7 \\
\ncffn{} (mult., bounded, $+$complement) & \textbf{8} & \textbf{8} & 7 & \textbf{8} \\
\bottomrule
\end{tabular}
\end{table}

\paragraph{Bounding buys width-efficient, shallow reasoning.}
At a single layer, the bounded multiplicative units are the most parameter-efficient reasoning basis by
a wide margin: \ncffn{} reaches $N{=}8$ and sigmoid-bilinear $N{=}7$, against GELU's $6$ and
raw-bilinear's $2$. A degree-2 layer cannot solve parity-8 by algebraic composition; the bounded units
do it through \emph{width}---combining sigmoid-bounded product terms in the read-out---and strikingly few
suffice. Sweeping the width at fixed $L{=}1$ (Table~\ref{tab:width}), each arm sits at a flat,
\emph{width-robust} shallow ceiling that is already reached by $d_{\mathit{ff}}{=}16$ and unchanged out to
$256$: \ncffn{} $8$, sigmoid-bilinear $7$, GELU $6$, and raw-bilinear below $5$ at every width. This is
exactly where bounding matters: products of $[0,1]$ values remain in $[0,1]$, so the read-out can stably
combine them, whereas an unbounded raw product explodes under summation---which is why raw-bilinear is
useless at one layer \emph{at any width}. Bounding is thus not a legibility nicety; it is what makes a
multiplicative unit a usable shallow reasoning basis.

\begin{table}[t]
\centering
\caption{Reach versus hidden width at fixed depth $L{=}1$ (mean over 4 seeds). The shallow ordering is
width-robust---each arm sits at a flat ceiling from $d_{\mathit{ff}}{=}16$ to $256$. Raw-bilinear does not
reach $N{=}5$ at any width (it solves only $N\!\le\!2$).}
\label{tab:width}
\begin{tabular}{lccccc}
\toprule
FFN type & $d_{\mathit{ff}}{=}16$ & $32$ & $64$ & $128$ & $256$ \\
\midrule
GELU             & 6 & 6 & 6 & 7 & 6 \\
Raw-bilinear     & $<5$ & $<5$ & $<5$ & $<5$ & $<5$ \\
Sigmoid-bilinear & 7 & 7 & 7 & 7 & 7 \\
\ncffn{}         & \textbf{8} & \textbf{8} & \textbf{8} & \textbf{8} & \textbf{8} \\
\bottomrule
\end{tabular}
\end{table}

\paragraph{Unbounding buys depth-composition.}
Raw-bilinear shows the opposite profile: its reach climbs $2\!\to\!4\!\to\!7\!\to\!8$ with depth,
tracking the algebraic fact that composing products doubles their degree per layer ($2^L$). Unbounded
products cannot be summed in width but \emph{can} be multiplied in depth, so depth is their only route
to parity---and a clean one. The bounded units do not gain comparably from depth (they are already near
ceiling at $L{=}1$, and the sigmoid squashes degree growth across layers), so the two strategies are
genuinely complementary: bounding trades depth-composition for width-stability.

\paragraph{The additive baseline plateaus.}
GELU sits between the two at every depth ($6$--$7$), neither width-efficient nor depth-composing for
parity---consistent with parity being additive-hard at any depth. Its near-flatness across $L$ confirms
that, unlike the multiplicative units, it derives no compositional benefit from added layers here.

\subsection{The complement adds a shallow margin}
The two bounded arms differ only in the explicit complement, and \ncffn{} edges sigmoid-bilinear where
depth cannot be leaned on: at $L{=}1$, reach $8$ versus $7$. With the operand budget tightest---one
layer---getting both $A\!\cdot\!B$ and $A\!\cdot\!(1-B)$ from a single shared operand pair, rather than
spending a separate pair to recover the complement by sign-flip, is worth a parity bit. The margin is
modest and concentrated at shallow depth; we do not claim the complement adds reasoning power beyond the
bounded multiplicative basis it shares with sigmoid-bilinear---its larger contribution is to legibility
(Section~\ref{sec:interp}). The honest summary is that \emph{multiplicativity} and \emph{bounding} are
the load-bearing capability properties, and the explicit negation is a small efficiency-and-legibility
addition on top.

\paragraph{Scope.}
These are controlled measurements on a single synthetic family with a fixed optimization budget;
the absolute reach values depend on width, steps, and seeds, and a longer budget would likely let
raw-bilinear's degree-$2^L$ advantage carry it higher at $L{=}4$. What is robust, and what we rely on,
is the \emph{qualitative dissociation}---bounded-shallow versus unbounded-deep, with the additive
baseline between---which holds across the sweep.

\section{Results II: Language Modeling}
\label{sec:lm}

We now place \ncffn{} in its target setting---a 125M-parameter decoder trained on OpenWebText---and ask
the necessary viability question: is an explicit-combination FFN a faithful parameter-neutral drop-in,
and what does the constraint cost? The answer is a clean tie on perplexity with a small, well-localized
cost on grammar, and---importantly---no language-model \emph{benefit}, exactly as the capability result
predicts.

\subsection{A perplexity tie that tightens, and no semantic gain}
At equal parameters, \ncffn{} matches the GELU baseline on perplexity, and the match tightens with
training (Table~\ref{tab:lm}): the dev-perplexity gap falls from $+0.11$ after one epoch to $+0.04$
after two, as the deep Boolean blocks relax toward GELU-equivalent capacity
(Section~\ref{sec:interp}). On the discriminating downstream benchmarks the model is at parity or
slightly behind. A single-seed epoch-1 LAMBADA edge (accuracy $+0.009$, perplexity $-7$) does
\emph{not} survive a second epoch---by epoch two the baseline is marginally ahead on both---so we read
LAMBADA as a tie and make no semantic-advantage claim. The floor benchmarks (HellaSwag, WinoGrande) are
at chance for both models at this scale.

\begin{table}[t]
\centering
\caption{Language-model evaluation, 12L, parameter-matched, across two epochs. The perplexity tie
\emph{tightens} with training; LAMBADA is a tie (the epoch-1 edge washes out); the BLiMP deficit
\emph{persists and widens}.}
\label{tab:lm}
\begin{tabular}{lcccc}
\toprule
 & \multicolumn{2}{c}{Epoch 1} & \multicolumn{2}{c}{Epoch 2} \\
\cmidrule(lr){2-3}\cmidrule(lr){4-5}
Metric & \textsc{gelu} & \ncffn{} & \textsc{gelu} & \ncffn{} \\
\midrule
Dev perplexity $\downarrow$    & 19.00 & 19.11 & 18.32 & 18.36 \\
LAMBADA acc.\ $\uparrow$       & 0.254 & 0.262 & 0.275 & 0.273 \\
LAMBADA ppl.\ $\downarrow$     & 109.4 & 102.4 & 78.1  & 85.5  \\
BLiMP (67) $\uparrow$          & 0.807 & 0.788 & 0.823 & 0.790 \\
\bottomrule
\end{tabular}
\end{table}

\subsection{A small, persistent grammatical deficit---and where it lives}
The one robust downstream signal is a BLiMP deficit, and it \emph{widens} with training rather than
closing: the baseline's grammar improves over the second epoch ($0.807\!\to\!0.823$) while \ncffn{}'s
stays flat ($0.788\!\to\!0.790$), so the gap grows from $-0.019$ to $-0.033$ (well outside within-sample
error). The deficit is concentrated, not diffuse: across the 67 subtasks the baseline wins 36 to 22
(9 ties), with losses clustered in \emph{island/long-distance extraction} (\textit{left-branch-island}
$-0.31$, \textit{coordinate-structure} $-0.25$), \emph{NPI licensing} (\textit{npi-present-1} $-0.18$,
\textit{matrix-question-npi} $-0.14$), and \emph{quantifiers} ($-0.10$ to $-0.13$). There is an irony
worth stating plainly: NPI licensing---the grammatical reflex of negation and polarity---is among the
negation-motivated architecture's \emph{weakest} areas (lone exception:
\textit{only-npi-licensor-present}, its largest gain at $+0.23$). The architectural negation primitive
does not translate into better handling of grammatical negation.

\subsection{Why: capacity allocation, not missing capability}
We attribute the deficit to capacity \emph{allocation} rather than an inability. Per parameter, the
hidden units cost the same as GELU's ($2d$ each), but a Boolean unit is a \emph{more constrained}
basis. A pair's two outputs satisfy $A\!\cdot\!B + A\!\cdot\!(1-B) = A$, so the pair spans only
$\{A,\,A\!\cdot\!B\}$: one standalone sigmoid feature and one product, with the second operand spent
entirely inside the product. Two GELU units of the same parameter cost instead give two \emph{independent}
standalone features. Ordinary language modeling is hungry for standalone, threshold-like features
(which is what grammar needs), and mostly does not need the products---so a quarter of \ncffn{}'s
budget is, for this task, spent on units of lower marginal value. Its effective standalone-feature
capacity is accordingly between that of a strictly narrower GELU model and the full baseline, which is
the deficit's size and sign. This is the same allocation that the capability probe rewards: the products
that cost grammar a little are exactly what reason on parity. The architecture's capability is real but
\emph{dormant} for next-token prediction, which does not call for it---so it ties, at a small constant
cost, rather than winning or losing decisively.

\section{Results III: Legibility and Its Dynamics}
\label{sec:interp}

Because every \ncffn{} unit is a named \textsc{and}/\textsc{and-not} of two operands, the layer admits
questions an opaque activation does not: \emph{where} in the network is genuine two-operand logic
realized, \emph{which} units are causal, and \emph{how} does the logical content change over training?
We answer these with inference-only probes (Section~\ref{sec:methodology}) and close with the result we
find most telling---that the amount of logic is not a property of the architecture but of what the task
rewards.

\subsection{Engagement is depth-uniform; realization is localized to layer 0}
The Boolean block is heavily and uniformly \emph{used}: its share of the per-token FFN residual write is
$0.30$--$0.47$ at every layer, with no decay toward the top (Table~\ref{tab:depth}). But genuine
\emph{two-operand} realization is sharply localized. Only at layer 0 is the second operand informative
($\bar B=0.48$, away from the uninformative $0.5$) with a sizable intersection signal
($\overline{A\!\cdot\!B}=0.13$); at every deeper layer $\bar B$ collapses to $\approx 0.50$ and
$\overline{A\!\cdot\!B}$ drops by an order of magnitude, so the units have degenerated to the
single-operand gates of Section~\ref{sec:degeneration}. We summarize this as a
\emph{recruitment-versus-realization} gap: the partition is recruited everywhere but its set-operator
semantics are realized only at the embedding-adjacent layer---a contrarian result, cutting against the
intuition that depth sharpens abstraction.

\begin{table}[t]
\centering
\caption{Boolean-block statistics by layer (final checkpoint). \emph{bool share}: fraction of the FFN
residual write carried by the Boolean block (engagement, depth-uniform). $\overline{A\!\cdot\!B}$,
$\bar A$, $\bar B$: operand realization---only layer 0 shows informative two-operand structure.}
\label{tab:depth}
\begin{tabular}{lcccccccc}
\toprule
Layer & 0 & 1 & 2 & 4 & 6 & 8 & 10 & 11 \\
\midrule
bool share              & 0.42 & 0.31 & 0.39 & 0.47 & 0.42 & 0.44 & 0.47 & 0.30 \\
$\overline{A\!\cdot\!B}$& \textbf{0.134} & 0.017 & 0.015 & 0.040 & 0.014 & 0.025 & 0.013 & 0.034 \\
$\bar A$                & \textbf{0.286} & 0.032 & 0.028 & 0.076 & 0.030 & 0.051 & 0.027 & 0.067 \\
$\bar B$                & \textbf{0.484} & 0.504 & 0.501 & 0.504 & 0.506 & 0.503 & 0.508 & 0.506 \\
\bottomrule
\end{tabular}
\end{table}

\subsection{The split is used or cleanly retired, almost never redundant}
\label{sec:operand}
A split representation invites a specific failure---the two operands collapsing into a \emph{redundant}
copy, so the layer pays for a duplicate. Section~\ref{sec:degeneration} predicts a different failure mode
instead: an unused unit drops \emph{one} operand and reverts to an ordinary single-operand activation. We
test this directly, classifying every Boolean unit from the joint statistics of its operands across
tokens. An operand is \emph{collapsed} when its activation variance is negligible; a surviving pair is
\emph{redundant} when $B$ correlates with $A$ above the $99$th percentile of a matched cross-unit floor,
and \emph{independent} otherwise. The three fates are exhaustive and consistent across the quantifier
ladder (Table~\ref{tab:operand}). \emph{Redundant} duplicate pairs are \textbf{under $5\%$} in every
variant, with median operand correlation $\approx 0$ (frequently negative---the ``$A$ and not $B$''
regime). The remainder divides into units that retire one operand to the single-operand gate of
Section~\ref{sec:degeneration} ($\approx\!45\%$) and units whose operands stay both active and independent
($\approx\!50\%$). The split therefore does not waste capacity on a duplicate: where two-operand logic is
not used, a unit relaxes to a plain activation, exactly the graceful degeneration the bounding affords.
We stress that ``independent'' is a far weaker condition than the \emph{informative} two-operand signal of
Table~\ref{tab:depth} (concentrated at layer~0) or the crisp predicates of Table~\ref{tab:logicdist}: an
operand may vary around its uninformative midpoint, contributing distinct but small-magnitude modulation.
The only claim here is the one the architecture promises---that the second operand, when present, is not a
redundant copy of the first.

\begin{table}[t]
\centering
\caption{Operand fate over all Boolean units (epoch-one checkpoints), averaged across the $12$ layers.
\emph{One-operand}: one operand collapsed to $\approx$\,constant (the single-operand relaxation of
Section~\ref{sec:degeneration}). \emph{Independent}: both operands active, correlation at/below a matched
cross-unit floor (distinct, non-duplicated signal). \emph{Redundant}: both active but the second tracks
the first above the floor. The split is almost never a redundant copy.}
\label{tab:operand}
\begin{tabular}{lccc}
\toprule
Model & One-operand & Independent & Redundant \\
\midrule
\textsc{+decay+gate} & $42.9$ & $54.0$ & $\mathbf{3.2}$ \\
\textsc{+decay}      & $43.4$ & $54.8$ & $\mathbf{1.8}$ \\
\textsc{+quant}      & $48.6$ & $50.6$ & $\mathbf{0.9}$ \\
\bottomrule
\end{tabular}
\end{table}

\subsection{Causal ablation: layer-0 is irreplaceable, deep blocks are redundant capacity}
Activation statistics measure use, not necessity. Zeroing the Boolean output columns and measuring the
perplexity penalty on a fixed held-out batch gives the causal picture (Table~\ref{tab:ablation}).
Removing \emph{only layer 0}'s Boolean block costs $\approx 187$ perplexity ($+2.23$ nats)---the single
most load-bearing FFN component in the network, and the one layer doing genuine two-operand logic.
Deeper layers are individually near-removable ($1$--$6$ ppl each) but \emph{collectively} essential:
ablating all Boolean blocks costs $+5.16$ nats, well above the $\approx 3.5$ nats of the summed
single-layer penalties---a distributed, redundant Boolean pathway across depth. A matched control (an
equal-sized random $\GELU$-column ablation) grows nearly as steeply over training
($+1.11\!\to\!+4.63$), so most of the all-block growth is generic co-adaptation, not Boolean necessity;
what survives the control is that Boolean columns are more load-bearing than $\GELU$ columns at every
checkpoint, an advantage that narrows by the end---the deep blocks converging toward generic capacity,
the localized layer-0 block remaining decisive.

\begin{table}[t]
\centering
\caption{Causal ablation: loss penalty (nats) for zeroing Boolean output columns, paired on a fixed
$131$k-token batch. The $\GELU$-chunk control ablates an equal number of random $\GELU$ columns/layer.}
\label{tab:ablation}
\begin{tabular}{lccc}
\toprule
Ablation ($\Delta\mathrm{loss}$, nats) & 50k & 100k & final \\
\midrule
All Boolean blocks                   & $+3.20$ & $+3.74$ & $+5.16$ \\
\quad $\GELU$-chunk control          & $+1.11$ & $+1.62$ & $+4.63$ \\
Layer-0 Boolean only                 & $+2.08$ & $+2.13$ & $+2.23$ \\
\quad (as perplexity)                & $+220$  & $+210$  & $+187$ \\
\bottomrule
\end{tabular}
\end{table}

\subsection{Readable units, and a logic--fuzzy spectrum}
Requiring a selective $A$-gate and a genuinely two-valued $B$ yields a small set---fewer than ten strict
two-operand units in the whole 12-layer model, almost all at layer 0---that read as recognizable
predicates \emph{without} dictionary learning. The clearest (layer-0, $A$-firing rate $0.085$) is a
function/content discriminator: its $A$-gate fires for high-frequency function tokens
(``\texttt{,}'', ``\texttt{\textbackslash n}'', ``\texttt{ the}'', ``\texttt{ .}''); the $A\!\cap\!B$
branch keeps the function tokens and the $A\!\setminus\!B$ branch isolates content tokens
(``\texttt{ Antonio}'', ``\texttt{ Nintendo}'')---a clean ``frequent token \textsc{and not} content
word.'' Others encode sentence/line-boundary context and a code-syntax-versus-identifier split. But this
is a \emph{spectrum}, not a dichotomy (Table~\ref{tab:logicdist}): of 257 active layer-0 units, 221 have
negligible $B$-discrimination (soft single-operand gating), with a thin tail of crisp logic; a deep
layer has no tail at all. The directly-readable units are few and encode low-level structural
distinctions; the bulk of operands are distributed and polysemantic, exactly as superposition predicts
\citep{elhage2022superposition}. The architectural payoff is not universal monosemanticity but that
\emph{even the distributed units carry explicit logical form}---each a known \textsc{and}/\textsc{and-not}
---whereas a $\GELU$ or bilinear ensemble is distributed \emph{and} opaque in its operation.

\begin{table}[t]
\centering
\caption{The logic--fuzzy spectrum at layer 0. Units binned by $B$-discrimination when $A$ fires
($\approx 0$ = soft single-operand gating). A deep layer (6) has no logic tail.}
\label{tab:logicdist}
\begin{tabular}{lccccccc}
\toprule
$B$-disc.\ bin & $0$--$.02$ & $.02$--$.05$ & $.05$--$.10$ & $.10$--$.15$ & $.15$--$.20$ & $.20$--$.30$ & active \\
\midrule
Layer 0 & 221 & 14 & 14 & 5 & 2 & 1 & 257 \\
Layer 6 & 5   & 0  & 0  & 0 & 0 & 0 & 5 \\
\bottomrule
\end{tabular}
\end{table}

\subsection{Redundant units, causally-used features}
Are the readable units \emph{causal}? Single-unit interventions say no: ablating the function/content
unit's readout, or flipping its $B$ operand, changes next-token predictions negligibly ($<10^{-3}$
nats). This is redundancy, not irrelevance: training distributes each feature across many carriers, so
no single unit is necessary (consistent with the super-additive block ablation above). The correct test
is at the level of \emph{features}: ablating a feature \emph{direction} removes it from all carriers at
once. Projecting the function/content axis out of the layer-0 output costs $148\times$ the loss of a
random direction of equal norm, localized to function-word positions ($1.6\times$); the
sentence-boundary unit's own axis gives $228\times$ a random direction and $4.2\times$ localization.
The faithful statement is therefore not ``this unit caused token $X$'' but \textbf{``a unit's firing is
faithful evidence that a causally-used feature is active here''}---and, methodologically, that under a
redundant network one must ablate \emph{directions}, not units, or systematically understate importance.

\subsection{The logic is task-shaped: it hardens where the task pays, erodes where it does not}
\label{sec:dynamics}
The most revealing property is dynamic. The layer-0 two-operand signal does not sharpen with training;
it \emph{softens}---$\overline{A\!\cdot\!B}$ decays monotonically across both epochs
($0.195\!\to\!0.150\!\to\!0.134\!\to\!0.111\!\to\!0.096\!\to\!0.087$), as the products relax toward the
single-operand gating that language modeling rewards (Section~\ref{sec:lm}). Crisp two-operand logic is
an \emph{early}-training feature that erodes; what persists is the depth-localization, the engagement,
and feature-level causality (the sentence-boundary axis is still $442\times$ a random direction at
epoch two), not unit-level crispness.

\begin{figure}[t]
\centering
\includegraphics[width=0.49\linewidth]{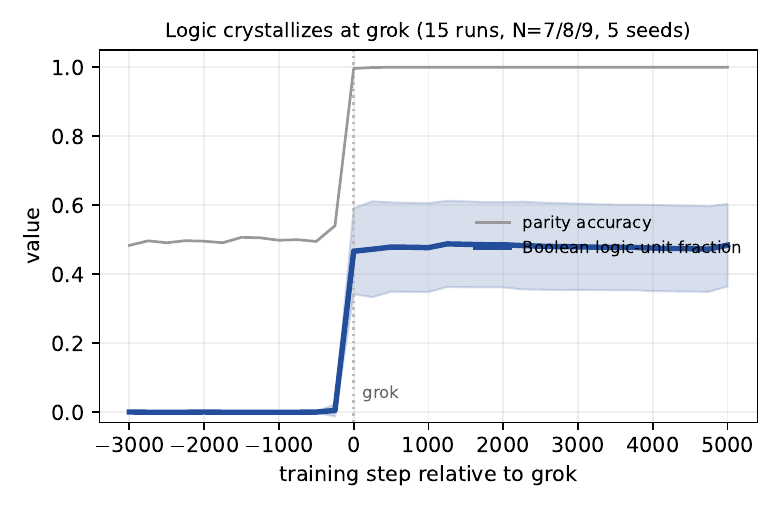}\hfill
\includegraphics[width=0.49\linewidth]{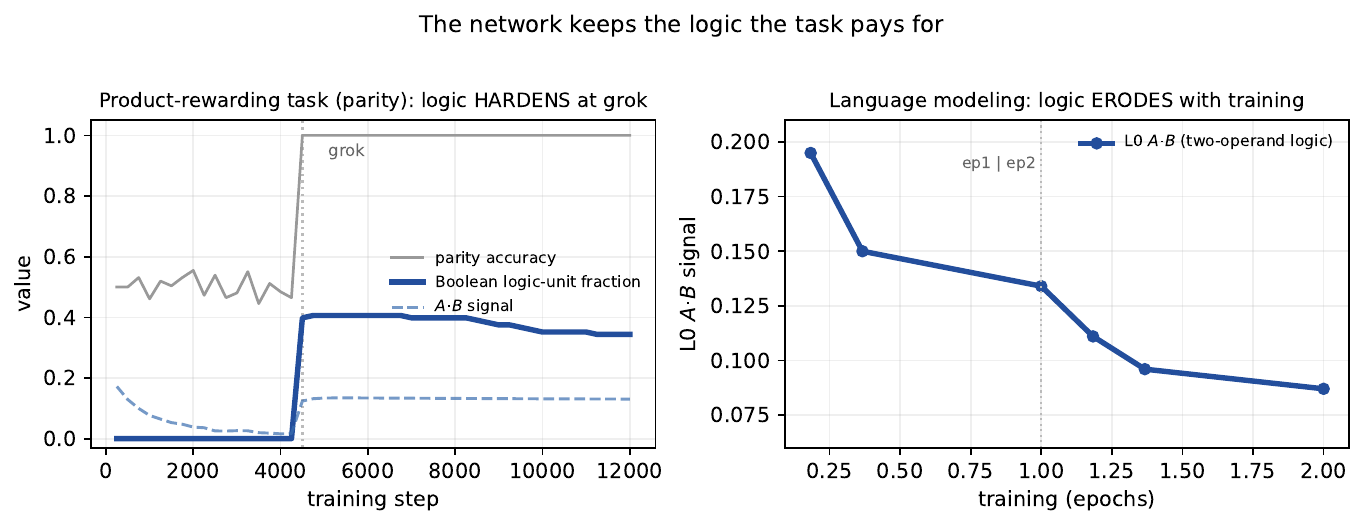}
\caption{The logical content is task-shaped. \textbf{Left:} on a product-rewarding task (parity), the
fraction of units doing two-operand logic snaps from $\approx 0$ to $\approx 0.48$ \emph{at} the grok
step (grok-aligned mean over 15 runs, band $=$ s.d.). \textbf{Right:} under language-model training, the
layer-0 two-operand signal \emph{erodes} monotonically across two epochs. Same Boolean block, same
probe, opposite trajectories.}
\label{fig:dynamics}
\end{figure}

Is the erosion intrinsic, or a readout of the objective? We test the converse directly. Train the
\emph{same} Boolean block, with the same probes, on a task that \emph{rewards} products---parity at the
solvable edge---and track the logic over training. It does the opposite: the Boolean structure
\emph{hardens}, and the hardening coincides exactly with generalization. Across 15 runs
($N\in\{7,8,9\}$, 5 seeds), grok-aligning each trajectory to its own accuracy jump, the fraction of
units doing genuine two-operand logic goes from $0.006$ \emph{before} the grok step to $0.475$
\emph{at} it---a step change in logical content locked to the moment the network acquires the capability
(Figure~\ref{fig:dynamics}, left). The contrast with the language-model erosion (right) is the result:
same architecture, same measurement, opposite trajectories. The network keeps exactly the logic its
objective pays for---crystallizing it at the instant of reasoning, relaxing it away where only
prediction is rewarded. ``How much logic'' is thus not a fixed attribute of the layer but a legible
readout of what the task demands of it.

\section{Results IV: A Trainability Ceiling at High Boolean Fraction}
\label{sec:trainability}

The architecture of Section~\ref{sec:capability} is a partition: a fraction of each FFN is the GELU
baseline and the remainder is the Boolean block, with the paper's configuration using a $25\%$ Boolean
share. A natural question for an \emph{end-to-end}-legible model is how far that share can be pushed---
ideally to $100\%$, a purely Boolean FFN with no GELU at all. We find a sharp obstacle that is not about
accuracy but about \emph{optimization}: as the Boolean fraction rises, the model trains normally for a
long time and then diverges abruptly, and the more Boolean the FFN, the sooner it does so.

\subsection{The trainable horizon shrinks as the Boolean fraction grows}
We trained the $12$-layer, $125$M-parameter model for one epoch on OpenWebText while sweeping the
Boolean fraction (all layers, $\{\cap,\setminus\}$ operators, parameter-matched throughout), and recorded
whether each run completed the epoch or diverged. We declare divergence when the \emph{instantaneous}
perplexity exceeds $100$ after a $15$k-step grace window, at which point training is halted automatically
and a restartable checkpoint is saved. Table~\ref{tab:trainability} reports the outcome.

\begin{table}[t]
\centering
\caption{Trainability versus Boolean fraction ($12$L, one epoch, all layers $\{\cap,\setminus\}$,
parameter-matched). Below $50\%$ the model completes the epoch; above it, every run diverges, and the
divergence step falls as the fraction rises. Single seed per configuration; the $90\%$/$95\%$ ordering is
within run-to-run noise.}
\label{tab:trainability}
\begin{tabular}{lcc}
\toprule
Boolean fraction & outcome & divergence step \\
\midrule
$25\%$ (this paper) & completes & --- \\
$50\%$              & completes & --- \\
$75\%$              & diverges  & $\sim$136k \\
$90\%$              & diverges  & $\sim$24k \\
$95\%$              & diverges  & $\sim$32k \\
$100\%$ (pure)      & diverges  & $\sim$16k \\
\bottomrule
\end{tabular}
\end{table}

The picture is a \emph{trainable-horizon} effect rather than a clean threshold: a $25\%$ or $50\%$ Boolean
model trains to convergence (the $50\%$ model reaching a dev perplexity of $21.6$ versus the GELU
baseline's $19.0$), whereas every model above $50\%$ eventually diverges, and the step at which it does so
grows shorter as the Boolean share grows---from $\sim$$136$k steps at $75\%$ down to roughly
sixteen thousand for the pure model. Crucially, the instability is also \emph{latent}: the $75\%$ model
trained healthily to step $123$k (perplexity $\approx 38$, indistinguishable from a stable run) before
diverging at $136$k, so a run that ``looks fine'' early is not yet safe. Even a single Boolean-bearing
layer in an otherwise-GELU stack is enough to seed the failure, though it postpones it dramatically (a
model with the Boolean block in layer~$0$ only diverged at step $202$k).

\subsection{The failure is a sudden spike, invisible to the running average}
The divergence is not a gradual drift but a near-instantaneous event. Across the $75\%$ model's blow-up,
the \emph{cumulative} epoch perplexity moved only from $92$ to $94$ while the \emph{instantaneous}
perplexity jumped from $51$ to $305$ within a single $250$-step logging window. The running average---the
quantity ordinarily monitored during training---barely registers the event until it is far advanced, which
is why our divergence criterion and the early-stopping guard both operate on instantaneous, not
cumulative, loss. We note this as a small methodological point: characterizing this failure mode at all
requires watching the instantaneous signal.

\subsection{It is not the magnitude of the residual write}
A natural hypothesis is that the Boolean block's contribution to the residual stream grows without bound
and a single large write destabilizes the model. The Boolean \emph{features} are already bounded---they
are sigmoid products in $[0,1]$---so the only unbounded quantity is the read-out matrix $W_o$. We tested
two interventions that bound the residual write directly. A parallel linear bypass added to a pure
Boolean layer delayed its divergence by roughly four-fold (step $16.5$k to $69$k) but did not prevent it;
normalizing the FFN output to unit RMS (a fixed-magnitude residual write regardless of $\|W_o\|$) likewise
only postponed the pure model's divergence (to step $19$k). \emph{Neither bounding the write nor a
stabilizing linear path removes the instability}, which argues against a write-magnitude explanation and
points instead to an ill-conditioning of the optimization itself.

A plausible---but unverified---mechanism is consistent with the interpretability measurements of
Section~\ref{sec:interp}. Training drives the operand projections well above their initialization scale
(the row norm of $W_a$ grows two- to four-fold over training) and pushes roughly half of the operand
values into the saturated tails of the sigmoid. In that regime the gradient of a bounded product
$\sigma(W_a x)\,\sigma(W_b x)$ is dominated by the small, sharply varying derivative of a saturated
sigmoid, and a stack of such layers is prone to the kind of sudden, self-reinforcing update we observe. A
sufficient minority of GELU units---the partition---supplies an unsaturating gradient path that keeps the
optimization well-conditioned, which is why a partition buys \emph{horizon} (more Boolean delays the
onset) without buying \emph{immunity}.

\subsection{Implication}
Legibility, in this architecture, is therefore free only up to a partition. At the $25\%$ Boolean share we
study, the model trains exactly as well as its GELU baseline; one can push to $50\%$ at a small perplexity
cost; but a fully Boolean, end-to-end-legible FFN is not trainable by the present recipe, and the two
obvious fixes---bounding the write, adding a linear highway---only delay the failure. The obstacle to an
explanatory language model in which \emph{every} feed-forward unit carries explicit logical form is thus
not representational but optimization-theoretic: the high-Boolean-fraction ceiling is consistent with a
saturated-product gradient that bounding the residual write does not cure. Removing it is the central open
problem this work leaves behind.

\section{Results V: Self-Forgetting Quantifiers and Legible Licensing}
\label{sec:quantifier}

Two results above bound \ncffn{}'s legibility, and they point the same way. Section~\ref{sec:interp}
found that genuine two-operand logic is realized only at the embedding-adjacent layer and \emph{erodes}
toward soft gating under language-model training. Section~\ref{sec:lm} traced the model's one robust
grammatical deficit to a specific cluster of constructions---NPI licensing, long-distance extraction,
and quantifiers---i.e.\ exactly the phenomena a set-logic FFN ought to be good at. Both diagnoses share a
cause: the operators of \ncffn{} are \emph{within-token}. Intersection and set-difference combine
features at the current position; they say nothing about whether a feature has \emph{already occurred}
earlier in the sequence, which is what licensing and quantification require (``there is a comparative
\emph{somewhere to the left}, so license \emph{than}'').

We close the paper by adding the missing primitive---a fuzzy \emph{quantifier} over the sequence---and
find that it converts the limitation into the work's strongest legibility result. A small block of
\emph{self-forgetting} fuzzy quantifiers (i) recovers the grammatical deficit and modestly exceeds the
baseline, (ii) \emph{holds} its logical structure through training---and pushes it \emph{into depth}
rather than eroding---and (iii) reads, at the semantic layers, as a bank of grammatical
\emph{licensing detectors}: crisp, short-lived, and individually nameable without dictionary learning.
The decay is the crux: it is what keeps the quantifier a \emph{predictive, local} operator rather than a
sticky latch, and what makes the mechanism fit how a transformer actually processes a sequence.

\subsection{A fuzzy-quantifier block with learned forgetting}
\label{sec:qblock}
To the hybrid layer of Section~\ref{sec:hybrid} (the $\rho=0.75$ $\GELU$/Boolean partition) we add a
third, parameter-neutral block of $m_q$ \emph{quantifier} units. Each unit forms a per-token fuzzy
membership from the residual stream,
\begin{equation}
M_t = \sigma(W_q\,x_t)\in[0,1]^{m_q},\qquad W_q\in\mathbb{R}^{m_q\times d},
\end{equation}
and the membership is aggregated \emph{along the sequence} by two causal fuzzy quantifiers, a
soft existential and a soft proportion, each with its own \emph{learned} forgetting rate:
\begin{align}
\text{(``$\exists$ recently'')}\quad & E_t = \max\!\big(M_t,\ \gamma\odot E_{t-1}\big), \label{eq:exists}\\
\text{(``proportion recently'')}\quad & P_t = (1-\lambda)\odot M_t + \lambda\odot P_{t-1}, \label{eq:prop}
\end{align}
with per-unit decays $\gamma=\sigma(\theta_\gamma)$ and $\lambda=\sigma(\theta_\lambda)$ in $(0,1)$. The
two limits are the non-forgetting quantifiers: $\gamma=1$ makes \eqref{eq:exists} a cumulative maximum
(``$M$ fired at \emph{some} earlier token''), and $\lambda=1$ makes \eqref{eq:prop} a cumulative mean
(``the \emph{fraction} of earlier tokens that fired''). We \emph{initialize} the decays near this
non-forgetting limit ($\gamma_0,\lambda_0\!\approx\!0.99$), so the block begins as a true sticky
quantifier and must \emph{learn} any forgetting it adopts. The aggregated streams $E_t,P_t$ are written
to the residual through the same shared, \emph{zero-initialized} read-out $W_o$ as the Boolean block, so
the layer again begins as a pure $\GELU$ FFN and wires the quantifiers in only if useful. We study three
members of a ladder, each over the same $\{\cap,\setminus\}$ Boolean base:
\textsc{+quant} (fixed $\gamma\!=\!\lambda\!=\!1$, no forgetting),
\textsc{+decay} (the learned $\gamma,\lambda$ above), and
\textsc{+decay+gate} (additionally a per-layer scalar gate $\beta=\sigma(\theta_\beta)$ on the quantifier
write, a learned per-depth volume control). Decay and gate parameters are per-unit/per-layer scalars and
are parameter-negligible; $m_q$ is chosen to hold the layer parameter-neutral with the $125$M baseline.
Logically, each unit is a \emph{unary} predicate $M$ quantified over position by
\eqref{eq:exists}--\eqref{eq:prop}: the block adds \emph{monadic} sequence quantification, not $n$-ary
relations, and the licensor-to-licensee dependency it captures is a relation between \emph{positions} bound
by the learned decay rather than an $n$-ary predicate (we return to this scope in
Section~\ref{sec:discussion}).

\subsection{The quantifier recovers the grammatical deficit---and ties LAMBADA on top}
\label{sec:qlm}
The motivating symptom was grammatical: \ncffn{} lost $-0.019$ BLiMP to the baseline, concentrated in
licensing (Section~\ref{sec:lm}). Adding the quantifier block \emph{closes that gap at epoch~1}
(Table~\ref{tab:qlm}). \textsc{+decay+gate} recovers BLiMP to $0.810$---level with the $0.807$
$\GELU$ baseline and well above \ncffn{}'s $0.788$---while simultaneously posting the strongest LAMBADA
of any model here ($0.271$ vs.\ the baseline's $0.254$), at unchanged parameters and on standard
attention. The recovery is \emph{monotone in the ladder}: the non-forgetting \textsc{+quant} only
\emph{partially} closes the gap ($0.795$, above \ncffn{}'s $0.788$ but short of baseline), \textsc{+decay}
reaches $0.800$, and \textsc{+decay+gate} $0.810$---so it is the forgetting, not the bare aggregation, that
closes the grammar. Epoch~2 sharpens this: the baseline's grammar pulls ahead ($0.823$), but the quantifier
\emph{halves} \ncffn{}'s now-wider deficit ($0.810$ vs.\ \ncffn{}'s $0.790$), and \textsc{+decay} posts the
best LAMBADA of any model here ($0.279$, edging the $0.275$ baseline---single seed, so we read it as a
tie-or-slight-lead). The mechanism we add to \emph{recover} the licensing deficit is the same one that,
below, reads as licensing detectors, and it is the \emph{decay} that makes the quantifier both grammatical
and legible.

\begin{table}[t]
\centering
\caption{Adding a sequence quantifier to \ncffn{} closes its grammatical deficit and leads the ladder on
LAMBADA ($12$L, parameter-matched, single seed, both epochs). At epoch~1 \textsc{+decay+gate} erases
\ncffn{}'s $-0.019$ BLiMP deficit ($0.810$, level with the $0.807$ baseline) and posts the best LAMBADA;
at epoch~2 the baseline's grammar pulls ahead ($0.823$) but the quantifier \emph{halves} the now-wider
\ncffn{} deficit, and \textsc{+decay} posts the best LAMBADA of any model here ($0.279$). The BLiMP
recovery is monotone in the decay ladder ($+$quant $<$ $+$decay $<$ $+$decay$+$gate).}
\label{tab:qlm}
\small\setlength{\tabcolsep}{4pt}
\begin{tabular}{lcccccc}
\toprule
 & \multicolumn{3}{c}{Epoch 1} & \multicolumn{3}{c}{Epoch 2} \\
\cmidrule(lr){2-4}\cmidrule(lr){5-7}
Model (12L, param-matched) & LAM acc.\ $\uparrow$ & LAM ppl.\ $\downarrow$ & BLiMP $\uparrow$ & LAM acc.\ $\uparrow$ & LAM ppl.\ $\downarrow$ & BLiMP $\uparrow$ \\
\midrule
\textsc{gelu} baseline                 & 0.254 & 109.4 & 0.807 & 0.275 & \textbf{78.1} & \textbf{0.823} \\
\ncffn{} ($\{\cap,\setminus\}$)        & 0.262 & 102.4 & 0.788 & 0.273 & 85.5 & 0.790 \\
\;\;$+$quant (no forgetting)           & 0.254 & 110.9 & 0.795 & 0.271 & 94.0 & 0.793 \\
\;\;$+$decay (learned $\gamma,\lambda$)& 0.250 & 110.3 & 0.800 & \textbf{0.279} & 87.6 & 0.800 \\
\;\;$+$decay$+$gate                    & \textbf{0.271} & \textbf{94.6} & \textbf{0.810} & 0.264 & 96.3 & 0.810 \\
\bottomrule
\end{tabular}
\end{table}

We read this conservatively---single seed, though now consistent across both epochs---but the qualitative
pattern is clean and is corroborated mechanistically below: the FFN block whose \emph{job} is sequence
quantification removes the deficit on the grammar that needs sequence quantification. Confirming that the
recovery is concentrated in the same licensing subtasks \ncffn{} lost (rather than diffuse) is the natural
follow-up; here we establish the aggregate recovery and the mechanism behind it.

\subsection{Now the structure holds---and migrates into depth}
\label{sec:qholds}
\begin{figure}[t]
\centering
\includegraphics[width=0.85\columnwidth]{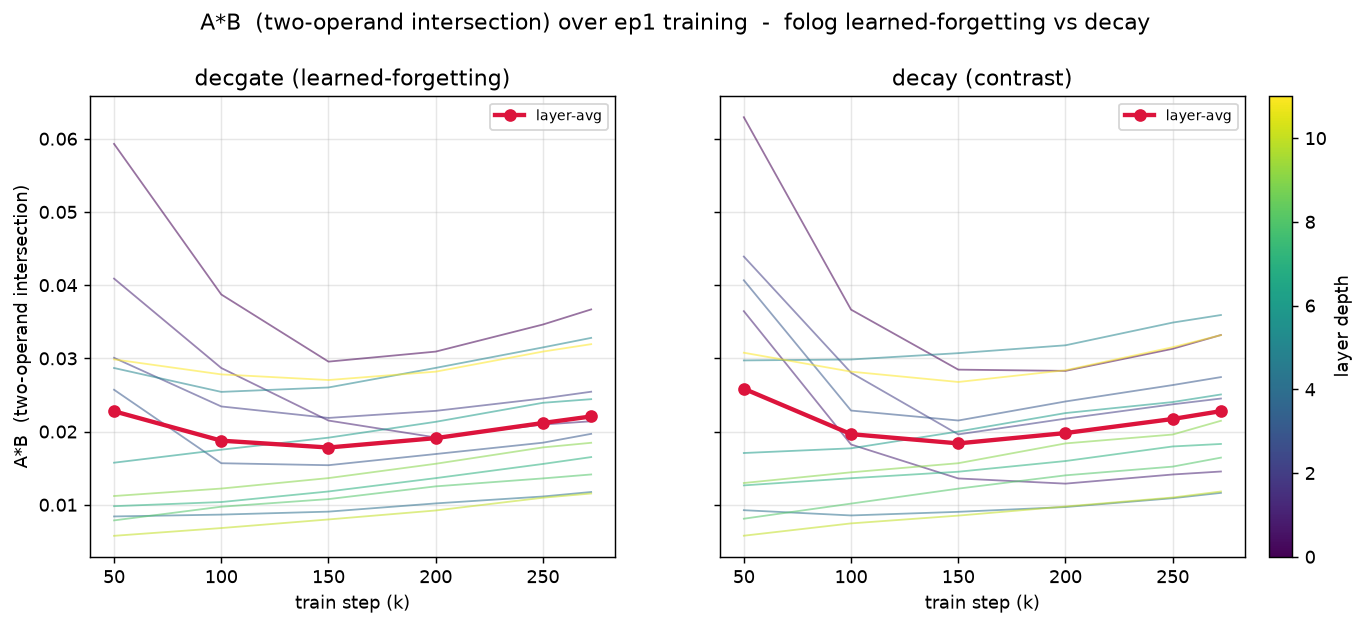}
\caption{$A\!\cdot\!B$ intersection magnitude per layer across epoch-1 training
(\textsc{+decay+gate}). Unlike \ncffn{}'s $\text{L}0$ erosion, the intersection signal migrates out of
layer~0 into the deep layers rather than collapsing---genuine two-operand logic is held and pushed into
depth.}
\label{fig:boolean-depth}
\end{figure}

The most worrying property of \ncffn{} was dynamic: its two-operand logic was an \emph{early}-training
feature that eroded toward soft gating (Section~\ref{sec:dynamics}). The quantifier model does not erode.
Re-running the realization probe across the same training checkpoints
($50\text{k}\to250\text{k}\to$ epoch) tells the opposite story (Table~\ref{tab:qdepth}). The load-bearing
intersection signal $\overline{A\!\cdot\!B}$ does not collapse to layer~0 and decay; it \emph{redistributes
toward depth}---falling at layer~0 ($0.059\to0.037$) while the deep layers roughly \emph{double}
($\text{L8}\;0.008\to0.014$, $\text{L9}\;0.011\to0.018$), with the layer-average tracing a shallow
U (a mid-training dip that recovers to its starting value by the epoch). Two surface statistics do soften
over training ($\mathrm{std}(B)\;0.26\to0.20$, $\mathrm{crisp}(A,B)\;0.53\to0.48$), but the operative
quantity---realized two-operand intersection---is held and \emph{deepened}. This is the reversal we were
after: where the within-token Boolean logic of \ncffn{} degenerated to an $\text{L}0$ relic, the
sequence-quantifier model spreads genuine logic across the stack and keeps it there
(Figure~\ref{fig:boolean-depth}).

The decay/gate split also separates here, and it is informative. \emph{Learned decay} preserves quantifier
crispness best (shallow layers even sharpen; only the top two or three soften, mean $0.858\to0.832$). The
\emph{gate} (\textsc{+decay+gate}) trades some of that away at the deepest layers ($0.840\to0.755$;
$\text{L}11$ $0.83\to0.51$) in exchange for its LAMBADA/BLiMP gains. We accordingly treat \textsc{+decay}
as the clean mechanistic object and the gate as a performance-oriented depth-shaping refinement.

\begin{table}[t]
\centering
\caption{Realization migrates into depth (\textsc{+decay+gate}, $\overline{A\!\cdot\!B}$ by layer over
training). Contrast Section~\ref{sec:interp}, where \ncffn{}'s two-operand signal was an $\text{L}0$-only
feature that decayed: here it falls at $\text{L}0$ and \emph{rises} in the deep layers.}
\label{tab:qdepth}
\begin{tabular}{lcccc}
\toprule
$\overline{A\!\cdot\!B}$ & L0 & L8 & L9 & layer-avg \\
\midrule
step 50k    & 0.059 & 0.008 & 0.011 & U-shape \\
epoch end   & 0.037 & 0.014 & 0.018 & (dips $\sim$150k, recovers) \\
\bottomrule
\end{tabular}
\end{table}

\subsection{The decay un-learns the latch: forgetting, not memory}
\label{sec:qdecay}
\begin{figure}[t]
\centering
\includegraphics[width=0.85\columnwidth]{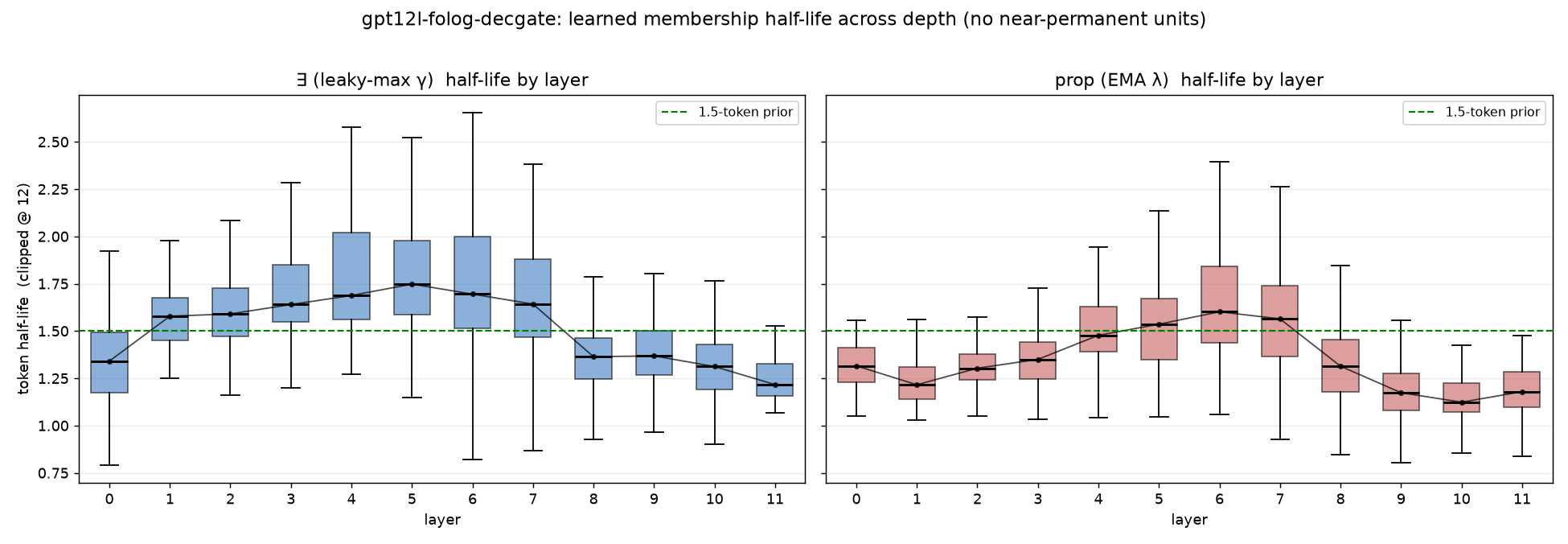}
\caption{Learned existential ($\exists$) token half-life per layer (median $\sim$$1.5$ tokens). From a
sticky $\gamma_0\!\approx\!0.99$ initialization, every unit learned a short, local memory; \emph{zero}
units remain near-permanent.}
\label{fig:halflife}
\end{figure}

The quantifiers \eqref{eq:exists}--\eqref{eq:prop} \emph{start} as sticky operators
($\gamma_0\!\approx\!0.99$: a cumulative max never forgets). A model that simply rode that initialization
would be implementing a permanent latch---``once true, true forever''---which is not how a language model
processes a sequence and would make the operator a poor fit. It does not. Reading the learned decays off
the trained weights and converting them to token half-lives $t_{1/2}=\ln(0.5)/\ln\gamma$ shows that
\emph{every} unit moved decisively away from the latch (Table~\ref{tab:qhalflife}):
\begin{itemize}
\item \textbf{No near-permanent units.} Not one of the $1536$ existential units (and none of the
proportional units) has $\gamma>0.97$ in either learned model; the maximum is $\gamma=0.963$, and only
$0.07\%$ exceed $\gamma=0.90$. From a $\gamma_0\!\approx\!0.99$ start, the decays were \emph{learned down}.
\item \textbf{Short, local memory.} The median existential half-life is $1.52$ tokens (proportional
$1.32$); $88.5\%$ of existential units forget within two tokens. The quantifier became a short recency
detector, not a store.
\item \textbf{A mid-stack lifetime hump.} Half-life peaks at the middle layers (L4--7, $\sim$$1.7$ tokens)
and is shortest at the edges (L0, L8--11, $\sim$$1.2$--$1.4$); the per-layer gate $\beta$ is
\emph{highest} exactly where lifetimes are shortest (peak $\beta=0.80$ at L9), i.e.\ the model leans on
the quantifier most for ephemeral, per-token detection.
\end{itemize}
That the operator un-learns its own stickiness is the result that makes it transformer-appropriate, and it
sets up the reading of the next subsection: a $\sim$$1.5$-token memory is precisely long enough to carry a
\emph{prediction} from one token to the next, and no longer (Figure~\ref{fig:halflife}).

\begin{table}[t]
\centering
\caption{Learned forgetting (existential quantifier, both learned models). Initialized at a sticky
$\gamma_0\!\approx\!0.99$ (cumulative max), every unit learned a short half-life; \emph{zero} units remain
near-permanent. Half-lives in tokens, $t_{1/2}=\ln(0.5)/\ln\gamma$.}
\label{tab:qhalflife}
\begin{tabular}{lcccc}
\toprule
 & median $t_{1/2}$ & \% units $<2$ tok & max $\gamma$ & \% units $\gamma>0.97$ \\
\midrule
existential ($E$) & 1.52 tok & 88.5\% & 0.963 & \textbf{0.0\%} \\
proportion ($P$)  & 1.32 tok & 96.4\% & 0.955 & \textbf{0.0\%} \\
\bottomrule
\end{tabular}
\end{table}

\subsection{The legible units are grammatical \emph{licensing detectors}}
\label{sec:qlicensing}
At the semantic layers where the quantifier write is largest (L9--L11, gate $\beta=0.80/0.75/0.56$), the
membership units are individually nameable, and they realize a single, consistent computation:
\emph{predictive grammatical licensing}. Crucially---and this is the cleanest part of the
story---a unit does \emph{not} fire on the token it decodes to. Reading a unit's write direction through
the unembedding (a logit lens over its $W_o$ column) names a \emph{function word}; but the membership
$M=\sigma(W_q x)$ itself fires one token \emph{earlier}, on the grammatical \emph{licensor} that predicts
that function word, and the short-memory existential then carries it forward to the position where the
function word is emitted. We verified the literal-token reading is false: for the comparative unit, mean
membership on `\,than' is $0.05$, but on its licensors `\,more'/`\,less' it is $0.55$--$0.64$. Four such
detectors recur across both learned models (Table~\ref{tab:detectors}):

\begin{table}[t]
\centering
\caption{Four grammatical \emph{licensing detectors} (layer~9, recurring across the \textsc{+decay} and
\textsc{+decay+gate} models). The membership fires on the \emph{licensor}; the unit writes toward the
\emph{licensed} function word it predicts; the $\sim$$1.5$-token decay bridges the gap. None fires on the
function word it names.}
\label{tab:detectors}
\begin{tabular}{llll}
\toprule
Construction & fires on (licensor) & writes / predicts & example (\,$M$ at licensor) \\
\midrule
comparison   & comparatives (`\,more', `\,rather', `\,-er') & `\,than' & ``rather'' $\to$ than ($0.95$) \\
passive/agentive & passive participles (`\,made', `\,caused') & `\,by' & ``made'' $\to$ by ($0.72$) \\
directional  & motion/transformation verbs (`\,turning', `\,put') & `\,into' & ``turning'' $\to$ into ($0.84$) \\
disjunction  & neg-polarity / alternatives (`\,neither', `\,either') & `\,or'/`\,nor' & ``neither'' $\to$ nor ($0.57$) \\
\bottomrule
\end{tabular}
\end{table}

\begin{figure*}[t]
\centering
\begin{subfigure}[t]{0.48\textwidth}
\centering
\includegraphics[width=\textwidth]{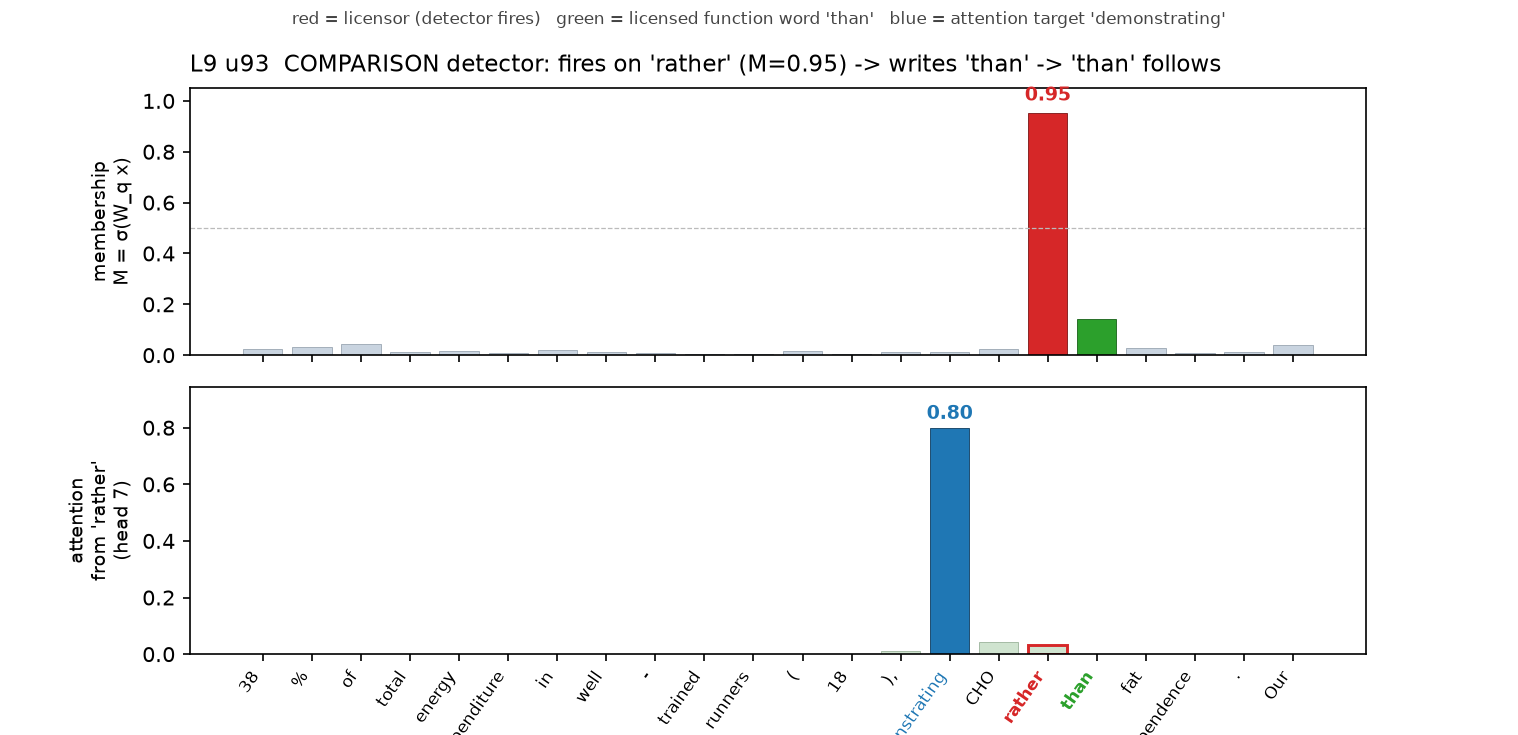}
\caption{Comparison detector: fires on comparatives (`\,rather'/`\,more') and predicts `\,than'.}
\label{fig:lic-comparison}
\end{subfigure}\hfill
\begin{subfigure}[t]{0.48\textwidth}
\centering
\includegraphics[width=\textwidth]{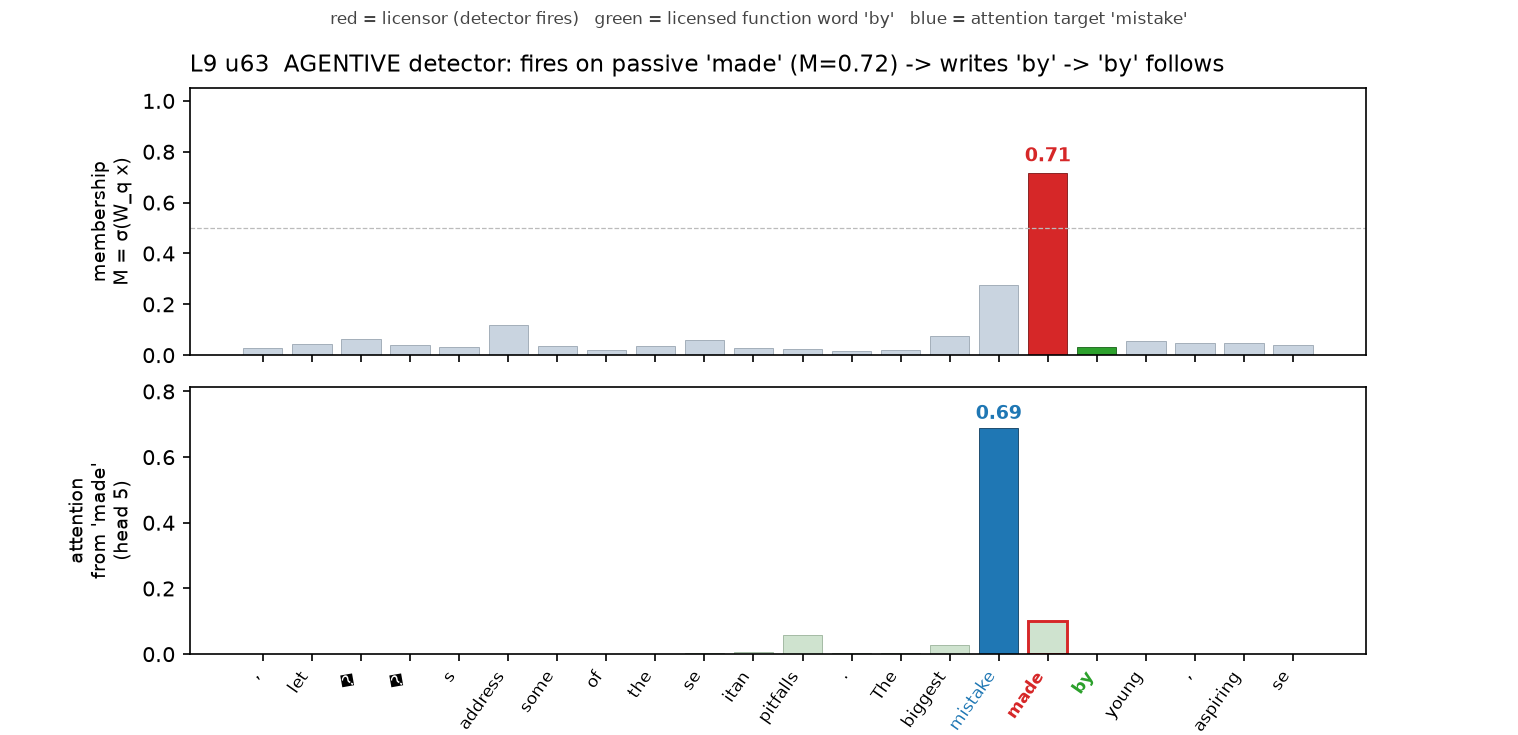}
\caption{Agentive detector: fires on passive participles (`\,made'/`\,caused') and predicts `\,by'.}
\label{fig:lic-agentive}
\end{subfigure}

\vspace{1em}

\begin{subfigure}[t]{0.48\textwidth}
\centering
\includegraphics[width=\textwidth]{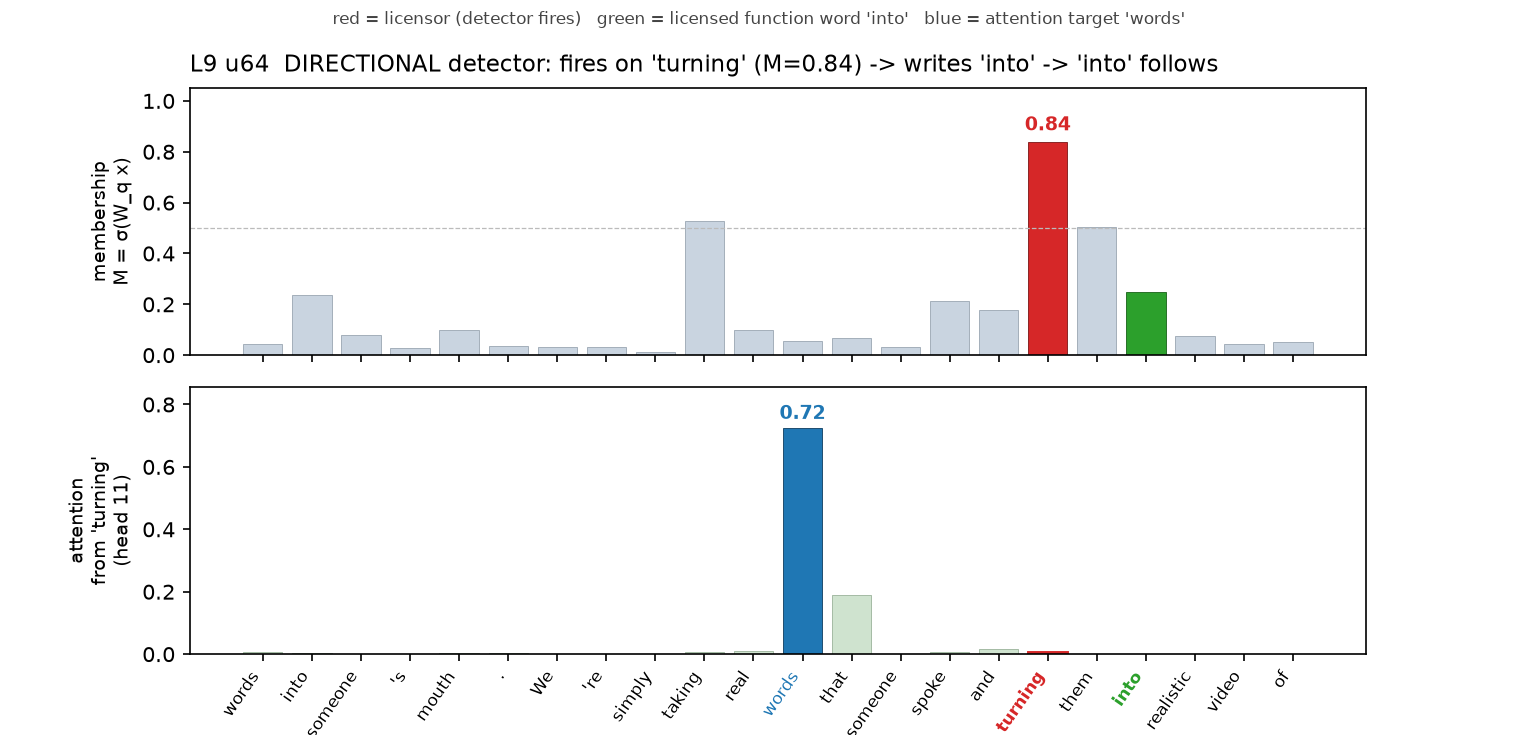}
\caption{Directional detector: fires on motion verbs (`\,turning'/`\,put') and predicts `\,into'.}
\label{fig:lic-directional}
\end{subfigure}\hfill
\begin{subfigure}[t]{0.48\textwidth}
\centering
\includegraphics[width=\textwidth]{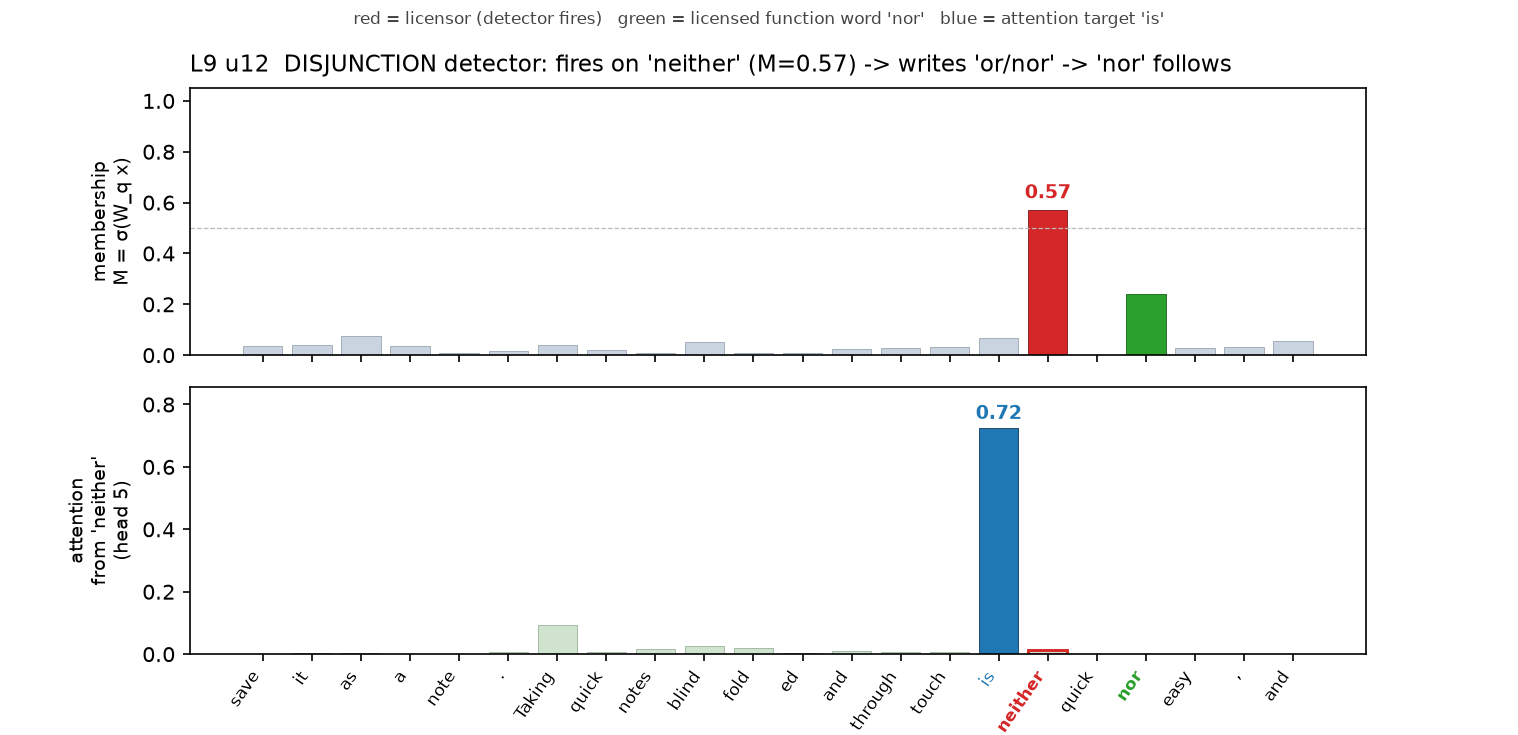}
\caption{Disjunction detector: fires on `\,neither'/`\,either' and predicts `\,or'/`\,nor'.}
\label{fig:lic-disjunction}
\end{subfigure}
\caption{Four grammatical \emph{licensing detectors} (layer~9), each a predictive operator. Membership
fires crisply on the grammatical \emph{licensor} (left spike), the $\sim$$1.5$-token existential decay
carries the signal forward over the licensing window, and the unit writes the \emph{licensed} function
word it predicts. None fires on the function word it names.}
\label{fig:licensing}
\end{figure*}

The mechanism is internally coherent in a way that is worth stating, because it explains why the decay was
necessary. In the four hand-inspected examples the licensor-to-function-word distance is one or two tokens
(``rather \underline{than}'' $+1$; ``turning them \underline{into}'' $+2$; ``neither quick \underline{nor}''
$+2$), and the learned existential half-life is $\sim$$1.5$ tokens. \emph{The decay tuned itself to the
grammatical-licensing window}: long enough to carry the membership from the licensor to the licensed
position, short enough not to latch and pollute later predictions. The three measurements of this
section---recovers the licensing deficit (\S\ref{sec:qlm}), holds and deepens its structure
(\S\ref{sec:qholds}), and forgets on a $\sim$$1.5$-token timescale (\S\ref{sec:qdecay})---thus compose into
one legible operator: \emph{fire on the licensor, decay for $\sim$one token, write the licensed word}
(Figure~\ref{fig:licensing}).
Each detector is, in effect, a learned, readable fragment of a transformer's grammatical competence---of
the kind the BLiMP suite tests directly.

\subsection{An honest ceiling: legibility is localized, not network-wide}
\label{sec:qceiling}
We resist over-reading the result. Quantifying \emph{how much of the whole network} is human-nameable---
combining, per layer, the token-explainability of the FFN units (Boolean and membership, value-vector to
vocabulary against a matched-random floor) with the discernibility of the attention heads (this model uses
standard attention; we score each head as crisp-and-nameable or not)---puts the figure at
$\approx\!2\%$ of all parameters, or $\approx\!4\%$ if one excludes the deliberate $\GELU$ gradient-highway
that carries no legibility claim. This is a logit-lens/single-relation concentration proxy, not a causal
autointerp score, and it should be read as a floor, not a verdict. The honest summary is that legibility is
\emph{localized}: FFN nameability is monotonically back-loaded, near the floor through L1--L7 and rising
sharply to L11 ($25.6\%$; membership alone $62.5\%$), exactly where the licensing detectors live; attention
nameability is the mirror image, present only as shallow positional heads (L2--L4, L7) with the deep half
opaque. So the claim of this section is precise and bounded: not an understandable network, but a
\emph{localized, interpretable-by-construction grammatical mechanism}---a bank of self-forgetting quantifier
units that, at the semantic layers, compute readable licensing---living inside an otherwise ordinary
transformer, at no parameter cost and no loss of language-model quality. That is still a sharp advance over
a standard FFN, whose interpretable-by-construction content is zero.

\section{Discussion}
\label{sec:discussion}

\paragraph{What the results say together.}
The thread connecting capability, language modeling, and legibility is a single claim: a more
\emph{structured} FFN reasons more compactly and exposes \emph{how} it computes, at a modest and
well-localized cost. Capability makes the ``compactly'' concrete---a single bounded-multiplicative
layer is a more parameter-efficient parity basis than strictly larger GELU
(Section~\ref{sec:capability}). Language modeling shows the cost is real but small and, tellingly,
that the capability is \emph{dormant} for next-token prediction, which does not reward products
(Section~\ref{sec:lm}). Legibility delivers the ``how''---explicit per-unit logical form, a directly
readable subset, causally-used features---and the dynamics tie it back to capability: the logic
\emph{is} the reasoning, crystallizing when the task rewards it and relaxing when it does not
(Section~\ref{sec:dynamics}).

\paragraph{The sequence operator turns the limitations into the headline.}
Two of those findings are limitations, and they share a cause: the set operators are \emph{within-token},
so they cannot express that a feature occurred earlier in the sequence (Section~\ref{sec:quantifier}).
Supplying the missing primitive---a fuzzy quantifier over the sequence with a \emph{learned} forgetting
rate---is the paper's strongest result, and it is striking for three reasons. First, the mechanism we add
to \emph{recover} the licensing deficit is the same one that, read out, \emph{is} the licensing: the block
whose job is sequence quantification both closes the grammar gap and exposes a bank of grammatical
licensing detectors. Second, the legibility no longer erodes---it holds and migrates into depth---so the
dynamic relaxation we documented for the within-token operators is specific to them, not to structured FFNs
in general. Third, and most telling for whether the operator belongs in a transformer at all, the decay
\emph{un-learns} its own stickiness: from a sticky cumulative-max initialization every unit learns a short,
$\sim$$1.5$-token half-life, and that half-life turns out to match the one-to-two-token distance from a
grammatical licensor to the word it licenses. The model was free to keep a permanent latch and instead
learned a local, predictive operator---exactly the temporal logic a left-to-right language model can
use---and the licensing detectors are what that operator looks like when it is read.

\paragraph{What kind of logic this is---and what it is not.}
It is worth stating the scope of the ``logic'' precisely, because the architecture invites a natural
misreading. The set operators are \emph{unary} predicates of a single token's representation, and
intersection and set-difference are \emph{propositional} connectives over them; the quantifiers
\eqref{eq:exists}--\eqref{eq:prop} range over the \emph{positions} of a unary predicate. The fragment is
therefore \emph{monadic} predicate logic---unary predicates with quantification over the single sequence
variable---\emph{not} $n$-ary relational logic, and we make no claim to relational expressiveness. This is
a scope, not a shortfall, for two reasons. First, monadic-with-sequence-quantification is already strictly
stronger than propositional: an existential over the prefix \emph{binds a variable}---``a feature occurred
at some earlier position''---which no within-token propositional layer can express, and it is exactly the
fragment a left-to-right model needs for the licensing dependencies we study. Second, the dependency the
quantifier captures---a licensor at one position to the word it licenses at another---is a relation between
\emph{positions}, bound by the decay-quantifier, not a relation between argument slots at one token: the
``second argument'' is the other position, and the learned decay is what binds it.
\ncffn{}'s operators are \emph{asymmetric} by construction
($A\setminus B\neq B\setminus A$; existential $\neq$ proportion), which is also why a unit's two operands do
not degenerate into a redundant representation: measured directly, redundant operand pairs are under $5\%$
of units, the rest either keeping independent operands or cleanly retiring one to a single operand
(Table~\ref{tab:operand}). We therefore keep the FFN's legible logic monadic and leave
relational binding to attention; the contribution is the legibility and the discovered licensing structure,
not expressiveness beyond the monadic fragment.
Explicit $n$-ary relational binding---role--filler structure rather than a single quantified variable---is a
different mechanism again, the province of tensor-product representations \citep{smolensky1990tpr} and the
relational reading of tensor logic \citep{domingos2025tensorlogic}; how to incorporate such constructions
effectively we leave to future work.

\paragraph{Multiplicativity, bounding, negation: a decomposition of the design.}
The capability probe lets us assign roles to the properties \ncffn{} adds over a GELU FFN.
\emph{Multiplicativity} is what does parity at all. \emph{Bounding} is what makes a multiplicative
basis usable in width (and so shallow), at the price of depth-composition---a genuine trade, not a
strict improvement, since unbounded products are the better deep basis. \emph{Negation} (the explicit
complement) adds a small shallow capability margin but is primarily a \emph{legibility} feature: it
turns the layer's products into named set operations and is what makes the readable units read as
\textsc{and}/\textsc{and-not} predicates. We are deliberate that the headline capability is a property of
the bounded multiplicative basis, shared with sigmoid-bilinear; the negation's distinctive contribution
is interpretive.

\paragraph{Limitations.}
The language-model results are single-seed at one scale (125M); the BLiMP deficit and the perplexity
tie are robust, but the finer downstream numbers want replication. The capability probe is a single
synthetic family (parity) under a fixed optimization budget; it measures a real property---per-parameter
reasoning efficiency---but a longer budget would let unbounded products' degree-$2^L$ advantage carry
further at depth, and other reasoning families may apportion the width/depth axes differently. And the
directly-legible units are a small minority; the bulk of operands are distributed, so \ncffn{} buys an
explicit combination \emph{rule} for free but not, in general, monosemantic operands. The quantifier
results (Section~\ref{sec:quantifier}) are likewise single-seed, though now reported across both epochs;
the aggregate grammatical recovery, the learned-forgetting statistics, and the licensing
detectors are clear, but a seed sweep and a per-subtask attribution of the BLiMP recovery remain to be run.
And the whole-network legibility figure we report is a logit-lens concentration proxy, not causal
autointerp, and it is localized to a few late layers---a readable mechanism inside the network, not a
readable network.

\paragraph{A trainability boundary, and a path to a fully-legible LM.}
The stabilizing $\GELU$ partition is required, not merely convenient. Section~\ref{sec:trainability}
characterizes a \emph{Boolean-fraction-dependent} trainability horizon: at or below an even split the
model trains a full epoch, but above it every run descends normally and then diverges abruptly---and the
more Boolean the layer, the sooner, from $\sim$$136$k steps at $75\%$ down to $\sim$$16$k for a fully
Boolean FFN. Tellingly, the two obvious remedies---normalizing the residual write to unit magnitude, and
adding a parallel linear highway---each only \emph{delay} the divergence rather than prevent it, which
points the cause at a saturated-product gradient pathology (the interaction of sigmoid saturation,
composition, and the residual stream) rather than the magnitude of the write. This is the chief obstacle
to an \emph{end-to-end}-legible language model, in which every FFN unit---not three-quarters of the
budget---is a named set operation. Having characterized the instability, we regard \emph{removing}
it---most plausibly through a remedy aimed at the saturated-product gradient rather than the residual
write---as the most promising next step: it is the difference between a legible \emph{component} and a
legible \emph{model}.

\paragraph{Where the capability should pay off.}
Because the capability is dormant for next-token prediction, the natural place to look for a
\emph{performance} win is tasks that reward multiplicative composition---symbolic, arithmetic, and
algorithmic reasoning---rather than open-web language modeling. The width/depth dissociation also
suggests a concrete design lever: bounded blocks where shallow, width-cheap logic is wanted; unbounded
(or hybrid) blocks where deep compositional reach is. We leave both to future work.

\section{Conclusion}
\label{sec:conclusion}
We studied a parameter-neutral transformer FFN whose hidden units are explicit fuzzy set operations,
with a bounded complement that gives negation a clean positive form. The choice is not free---it costs a
little grammar and needs a stabilizing partition---but it buys two things an opaque activation does not.
It reasons more compactly: on a controlled probe, bounded multiplicative units are the most
parameter-efficient reasoning basis at shallow depth, and the properties we toggle (multiplicativity,
bounding) cleanly govern the width and depth axes of that efficiency. And it is legible: every unit is a
named operation, a subset reads as predicates, and the network's logical content is a dynamic readout of
what its objective rewards,
crystallizing at the moment of reasoning and eroding where only prediction is. And when those within-token
operators are given an explicit \emph{sequence} companion---a fuzzy quantifier free to learn how long to
remember---the FFN's hardest grammatical failures become its most legible mechanism: a bank of
\emph{self-forgetting} quantifier units that, at the semantic layers, compute readable grammatical
\emph{licensing}, recover the grammar deficit at no parameter or quality cost, and each learn a memory just
long enough to carry a licensor to the word it licenses. That legibility is for now
free only up to a partition: a stabilizing $\GELU$ fraction is required, and training a fully Boolean,
end-to-end-legible model---removing the trainability boundary of Section~\ref{sec:trainability}---is the
open problem we leave.
A structured component, in short, lets us ask not just what a feed-forward layer represents but how it
combines---and, with an explicit sequence operator, how it \emph{licenses}---and, increasingly, get an
answer.

\bibliographystyle{plainnat}
\bibliography{refs}

\end{document}